\documentclass[10pt]{article} % For LaTeX2e
\usepackage[accepted]{tmlr}
\usepackage{amsmath,amsfonts,bm}

\def\eqref#1{equation~\ref{#1}}
\def\1{\bm{1}}

\def\vx{{\bm{x}}}
\def\vy{{\bm{y}}}

\DeclareMathAlphabet{\mathsfit}{\encodingdefault}{\sfdefault}{m}{sl}
\SetMathAlphabet{\mathsfit}{bold}{\encodingdefault}{\sfdefault}{bx}{n}

\usepackage{hyperref}
\usepackage{cleveref}
\usepackage{url}
\usepackage{microtype}
\usepackage{graphicx}
\usepackage{subcaption}
\usepackage{booktabs} % for professional tables
\usepackage{wrapfig}
\usepackage{tcolorbox}
\usepackage{multirow}
\usepackage{floatrow}

\title{Reliability Scaling Laws for Quantized Large Language Models}

\author{\name Sirine Ayadi\thanks{Equal contribution.} \email sirine.ayadi@tum.de \\
      \addr School of Computation, Information and Technology, Technical University of Munich \\
      Munich Data Science Institute
      \AND
      \name Sándor Daróczi$^{*}$ \email 
sandor.daroczi@tum.de \\
      \addr School of Computation, Information and Technology, Technical University of Munich \\
      Munich Data Science Institute
      \AND
      \name Stephan Günnemann \email s.guennemann@tum.de\\
      \addr School of Computation, Information and Technology, Technical University of Munich \\
      Munich Data Science Institute \\
      Pruna AI
      \AND
      \name Bertrand Charpentier \email bertrand.charpentier@pruna.ai\\
      \addr Pruna AI}

\def\month{07}  % Insert correct month for camera-ready version
\def\year{2026} % Insert correct year for camera-ready version
\def\openreview{\url{https://openreview.net/forum?id=UUBijehMQO}} % Insert correct link to OpenReview for camera-ready version

\begin{document}

\maketitle

\begin{abstract}
\emph{Quantization} is a powerful strategy to build capable and resource-efficient large language models (LLMs) by reducing the bitwidth of the parameters. While quantized LLMs achieve state-of-the-art performance on unperturbed inputs using standard predictive metrics, their performance on perturbed inputs, measured using reliability metrics, remains underexplored, despite its importance for reliable deployment. To address this gap, we {first} conduct a comprehensive reliability evaluation of quantized LLMs consisting of three key components: \textbf{(1)} \emph{Uncertainty:} We assess the trustworthiness of LLMs quantized to 2, 3, 4, and 8 bits using six different quantization methods, employing established uncertainty metrics. {\textbf{(2)} \emph{Calibration:} We assess how well-calibrated the uncertainty estimates of quantized models are across model scales and bit precisions.}
\textbf{(3)} \emph{Robustness:} We design character-level and word-level input perturbations to evaluate the reliability of quantized models under semantically-preserving variations in the inputs that arise in real-world applications.
{Second, we characterize how reliability scales with the total number of model bits. }
%\textbf{(3)} \emph{Reliability scaling trends:} We investigate how the reliability scales with the number of model bits. 
Our study reveals that while the performance scales monotonically with the total number of bits, the reliability scalings are nonlinear. A reliability peak occurs for 4-bit quantized models, indicating that quantizing moderately sized models offers the best reliability-efficiency trade-off. Additionally, our empirical findings reveal that quantization enhances the robustness of LLMs to natural input perturbations. 
\end{abstract}

\section{Introduction}\label{sec:introduction}
Large language models (LLMs) \citep{touvron2023llama, team2024gemma,wei2022emergent} have reshaped the field of Natural Language Processing (NLP) with their remarkable performance across a range of complex benchmarks. Recent advances in LLMs are based on the principle that model performance improves predictably with increased model size and training data, following well-characterized scaling laws \citep{kaplan2020scaling}. However, their large size and high computational needs pose significant challenges for practical use, especially in resource-limited settings. This has spurred interest in model compression to reduce inference latency and memory requirements, including quantization \citep{dettmers_2022_bitsandbytes, lin2024awq, frantar2022gptq}, pruning \citep{sun2023simple, frantar2023sparsegpt}, and knowledge distillation \citep{gu2023minillm, liang2023less}.

Despite rapid progress in LLM quantization, their evaluation has predominantly focused on benign task performance, emphasizing that compressed models should retain the accuracy of the base model on downstream tasks \citep{lin2024awq, frantar2022gptq, sun2023simple, frantar2025compression}. While useful, these evaluations ignore critical reliability aspects, notably uncertainty and robustness, which are essential for a safe and trustworthy deployment. 
On the one hand, uncertainty quantification \citep{kadavath2022language} and calibration \citep{krishnan2024enhancing} have gained significant traction to assess the trustworthiness of responses generated by LLMs. However, the impact of model quantization on these critical dimensions remains underexplored. On the other hand, quantized models deployed in practice may encounter minor input noise, such as typos or grammatical errors, while the semantics of the original sentence are preserved \citep{moradi2021evaluating}. Yet, commonly used benchmarks \citep{joshi2017triviaqa, reddy2019coqa} for evaluating the capabilities of quantized models fail to capture such perturbations, limiting our understanding of model robustness in real-world applications.
%To evaluate the robustness to such perturbations, we implement character-level and word-level perturbations to perturb question input prompts, and assess the robustness of base and quantized models. A robust model should maintain consistent predictions under semantically-preserving variations in the input.

To address these gaps, we conduct a comprehensive evaluation of quantized LLMs across multiple \textbf{reliability} dimensions, including \emph{uncertainty}, and \emph{robustness} to semantically-preserving input perturbations. To better understand how many bits should be used to improve the reliability-efficiency trade-off, we leverage \textbf{bit-level scaling laws}, which reveal underlying trends that extend beyond individual data points. Consider two models: 
One model trained from scratch with 4 billion parameters at 16-bit precision, and another model with 16 billion parameters quantized to 4-bit precision. While both have the same total number of bits, their behavior can differ significantly \citep{dettmers2023case}. 
Prior work \citep{xu2024beyond, dutta2024accuracy,lin2024awq,dettmers2023case} mainly focus on assessing the effectiveness of the quantization methods by solely relying on the downstream task performance. In particular, \citet{dettmers2023case} explore bit-level scaling trends for standard task performance on different benchmarks, showing that accuracy typically improves with the total number of bits. 
However, we find that higher accuracy does not necessarily correspond to higher reliability.

\begin{figure}[htpb]
 \vspace{-0.5em}
    \centering
    \includegraphics[width=0.9\textwidth]{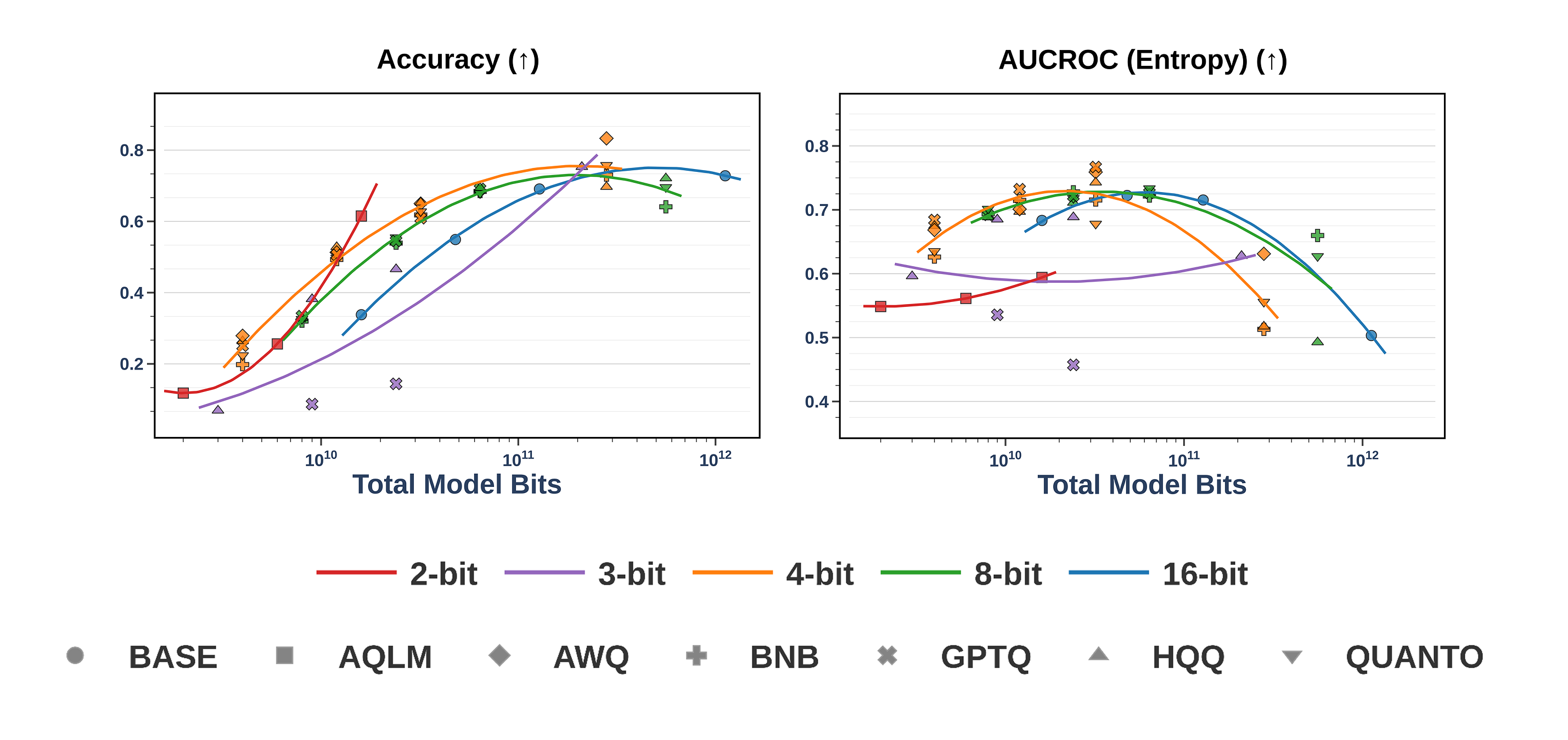}
    %{figures/intro_acc_entropy_triviaqa.png}
    \caption{Bit-level scaling trends of the accuracy and AUCROC(Entropy) on  TriviaQA. We use four base models (blue): LLaMA-3.2-1B, LLaMa-3.2-3B, LLaMa-3-8B, and LLaMa-3-70B, and {all available corresponding quantized variants} using six quantization methods and different bitwidths.} 
    \label{fig:intro_scaling_plots}
    \vspace{-0.5em}
\end{figure}

As shown in \cref{fig:intro_scaling_plots}, accuracy scales monotonically with the total number of model bits. In contrast, reliability follows a non-monotonic trend: reliability peaks for moderately-sized models quantized to 4 bits, suggesting that an optimal trade-off can be achieved without resorting to either high-precision quantization or the largest model.

We highlight our main contributions in the following:
\begin{itemize}

\item {We introduce a reliability evaluation framework for full-precision and quantized LLMs, consisting of three key components:} (i) predictive uncertainty, (ii) calibration, and (iii) robustness to semantically-preserving perturbations. To this end, we implement 15 natural character-level and word-level perturbations that commonly arise in typed digital communication.

%\item We study the \emph{reliability bit-level scaling laws} to examine the precision that maximizes the reliability for a fixed budget of total model bits. 

\item {We characterize reliability bit-level scaling trends to examine the precision that maximizes reliability for a fixed budget of total model bits.} We conduct the first comprehensive reliability evaluation of six state-of-the-art quantization techniques across models with 1 billion to 70 billion parameters. We examine whether reliability scaling trends are consistent across different model architectures and benchmarks. We find that 4-bit precision offers the best trade-off between reliability and efficiency across tasks, model families, and quantization methods.

\item We further investigate why 4-bit precision yields favorable performance, and examine whether the observed scaling trends persist under different LLM pruning strategies.

%\item We implement 15 character-level and word-level natural input perturbations, and study how quantization affects the robustness to semantically-preserving input noise.
%\item We study the bit-level scaling trends of post-training quantization methods to determine the precision that maximizes robustness given a specific budget of total model bits.

%\item We conduct the first comprehensive reliability evaluation of six state-of-the-art quantization techniques, on both \emph{unperturbed} and \emph{perturbed} input prompts. We examine which quantization methods improve the reliability scalings for 2 to 8-bit precision at scales of 1B to 70B parameters across LLaMA, OPT, and Qwen model families. Our key findings are: \emph{(i)} Base and quantized models are sensitive to different character-level and word-level input perturbations. Our empirical results demonstrate that quantization not only enhances efficiency but also improves the robustness of LLMs to semantically-preserving input perturbations. \emph{(ii)} Zero-shot performance scaling trends exhibit a different behavior than the reliability scaling trends.\emph{(iii)} 4-bit precision offers the best reliability-efficiency trade-off across different tasks, model families, and quantization methods.
\end{itemize}
\vspace{-0.5em}
\section{Related Work}\label{sec:related_work}
\vspace{-0.5em}
\paragraph{Scaling Laws for LLMs}
This work builds on established scaling laws that characterize how the training budget affects the performance of language models \citep{kaplan2020scaling, hernandez2021scaling, sorscher2022beyond, kumar2024scaling}.
\citet{kaplan2020scaling} demonstrate that the model loss follows a power law with the number of model parameters and tokens. Building on this, \citet{hoffmann2022training} argue that both model size and the number of training tokens should be scaled equally to achieve optimal compute training. Other works have shifted focus to the test-time budget, examining how quantization can impact the performance \citep{dettmers2023case, kumar2024scaling,frantar2025compression}. In this work, our goal is to understand the impact of the compression budget, defined as the total number of model bits, on downstream performance. Closest to our work are the bit-level scaling laws for zero-shot performance of quantized LLMs proposed in \citet{dettmers2023case}, where the goal is to determine the precision that maximizes the accuracy. However, \citet{dettmers2023case} solely focus on scaling trends of downstream task performance, and do not account for critical reliability dimensions necessary for safe deployment.

%\paragraph{Evaluating Quantized LLMs}
%Quantization is a widely used compression technique that reduces the number of bits (i.e., precision) in the parameters of a model with minimal loss in inference performance. We focus on Post-Training Quantization (PTQ) techniques \citep{dettmers_2022_bitsandbytes, lin2024awq, frantar_gptq_2023, badri_half-quadratic_nodate, badri_towards_nodate, optimumquanto_2023, egiazarian_extreme_2024, ashkboos_quarot_2024, lin_qserve_2024}, which quantize pre-trained models without re-training. To assess the performance of quantized LLMS, existing works \citep{lin2024awq, frantar2022gptq} primarily focus on preserving the perplexity or the accuracy of question-answering benchmarks. \citet{dettmers2023case} conduct the first study on bit-level inference scaling laws for quantization methods to determine the precision that maximizes the accuracy and reveal that $4$-bit methods yield an optimal zero-shot performance. While \citet{dettmers2023case} solely focus on the scaling behavior of the accuracy, we examine the scaling of safety-critical dimensions. \citet{hong2024decoding} suggest that solely relying on the perplexity is not sufficient and propose a comprehensive evaluation of compressed LLMs across various trustworthiness dimensions such as stereotype, toxicity, privacy, fairness, ethics, adversarial robustness, and out-of-distribution robustness. \citet{hong2024decoding} reveal that quantization is more effective than pruning as it achieves the best trade-off between efficiency and trustworthiness. 

\paragraph{Reliability Evaluation of LLMs}
With the increasing interest in LLMs, understanding uncertainty and calibration is crucial. Various uncertainty quantification methods have emerged to determine the trustworthiness of the generated responses by LLMs \citep{kuhn2023semantic, ye2024benchmarking}. \citet{kuhn2023semantic} introduce semantic entropy, which considers linguistic invariants arising from shared meanings. \citet{malinin2020uncertainty} introduce predictive entropy, which quantifies the uncertainty of auto-regressive models based on their own outputs. Recently, \citet{ye2024benchmarking} employed conformal prediction to quantify the uncertainty of LLMs in NLP tasks, including question answering and reading comprehension. Calibrating the uncertainty estimates is essential to evaluating the reliability of LLMs. A well-calibrated model correlates low uncertainty with accurate responses and high uncertainty with likely incorrect responses. This is typically evaluated using the Expected Calibration Error (ECE) \citep{jiang2021can} and the Brier Score \citep{brier1950verification}. In generative tasks, defining calibration is often challenging \citep{kapoor2024calibration}, especially for variable-length response sequences. 

While several studies have explored different reliability dimensions for LLMs, these aspects remain underexplored for quantized models. Most prior works on model compression, such as \citet{lin2024awq, frantar2022gptq, frantar2023sparsegpt, optimumquanto_2023, ashkboos_quarot_2024}, primarily evaluate the effectiveness of compression methods through standard performance metrics, including perplexity and accuracy on various benchmarks. However, these benchmarks fail to capture critical aspects of model reliability. Recently, \citet{hong2024decoding} present a comprehensive evaluation of compressed LLMs across various trustworthiness dimensions including stereotype, toxicity, privacy, fairness, ethics, adversarial robustness, and out-of-distribution robustness. Their primary question of interest is how to construct trustworthy 7B models, either by pre-training from scratch or by compressing a larger pre-trained 13B model.
\citep{hong2024decoding} reveal that quantization offers a more favorable trade-off between efficiency and trustworthiness than pruning.

\section{Reliability Evaluation Framework}\label{sec:evaluation_method} 
Consider a pre-trained autoregressive language model $P_{\theta}\left(\mathbf{y} \mid \mathbf{x} \right)$ parameterized by $\theta$, where $\mathbf{x}$ is an input prompt, and $\mathbf{y}$ is the generated sequence. We adapt this model to downstream conditional text generation tasks, such as question answering. Each task is represented by a dataset of context-target pairs, $\mathcal{D} = \{(\mathbf{x}_i, \mathbf{y}^{*}_i)\}_{i=1}^{N}$, where both $\mathbf{x}_i$ and
$\mathbf{y}^{*}_i$ are sequences of tokens. Given an input prompt $\mathbf{x}$, the model generates answer tokens sequentially as follows: 
\begin{equation}\label{eq:autoregressive-gen}
        P_{\theta}\left(\mathbf{y} \mid \mathbf{x} \right) = \prod_{t=1}^{T} P_{\theta}\left(\vy_t \mid \vy_{1}, \dots, \vy_{t-1}, \mathbf{x} \right) \:,
    \end{equation}
where $\mathbf{y} = (\vy_1, \dots, \vy_T)$ is the final generated sequence consisting of $T$ tokens. 
We denote the model's predicted token distribution by $P_{\theta}\left(\vy \mid \vy_{1}, \dots, \vy_{t-1}, \mathbf{x} \right)$, where $\vy \in \mathbb{R}^{\mid\mathcal{V}\mid}$, and $\mathcal{V}$ is the vocabulary. At each decoding step $t$, the model samples a token $\vy_t \sim P_{\theta}(. \mid \vy_{1:t-1}, \mathbf{x})$. 

To study the bit-level scaling, we consider quantization, which is a widely used compression technique that reduces the number of bits in the parameters of a model with minimal loss in inference performance \citep{dettmers_2022_bitsandbytes, lin2024awq, frantar2022gptq, optimumquanto_2023, badri_half-quadratic_nodate}.
\subsection{Reliability Evaluation}\label{sec:reliability_metrics}
In this paper, our goal is to evaluate various aspects of \emph{reliability} for quantized LLMs in a question answering setting. 
Given an input prompt $\mathbf{x}$, $P_{\theta}(\mathbf{y}\mid\mathbf{x})$ spans a probability distribution of all possible output sequences. 
Although we are generally interested in the output distribution $P_\theta(\mathbf{y} \mid \mathbf{x})$, in practice, we cannot directly access this distribution since the number of possible output sequences $|\mathcal{V}|^{T}$ is large. Therefore, we first define our reliability metrics at the \emph{token-level}, as we have direct access to the token distributions $P_{\theta}(\vy \mid \vy_{1:t-1}, \mathbf{x})$ at every step $t$, and aggregate these measures to obtain \emph{sequence-level} metrics.
Let $M_t$ denote a token-level metric at step $t$, the sequence-level metric is computed as the average across all time steps:
\begin{equation}
    M_\text{seq} = \frac{1}{T} \sum_{t=1}^{T} M_t \:.
\end{equation} 
Relying on a single generated sequence $\mathbf{y}$ given an input prompt $\mathbf{x}$ can lead to an incomplete assessment in a natural language generation (NLG) setting. To provide a more robust and representative estimate of the model's reliability, we sample multiple sequences per input prompt and report averaged results across these samples. 
%Additional details on the sampling are provided in \cref{app:evaluation_datasets}.
In the following, we outline the different reliability metrics measured on the token level.
%\paragraph{Entropy.} 
To measure the model's uncertainty in the predicted tokens, we compute the \emph{token-level entropy}, which corresponds to the Shannon entropy of the token distribution \citep{fomicheva2020unsupervised, malinin2020uncertainty}:
\begin{equation}\label{eq:token-entropy}
H_t= - \sum_{k=1}^{\mid\mathcal{V}\mid} P_{\theta}\left(\vy_{k} \mid \vy_{1:t-1}, \mathbf{x}\right) \log P_{\theta}\left(\vy_{k} \mid \vy_{1:t-1}, \mathbf{x}\right) \:,
\end{equation}
measured across the entire vocabulary $\mathcal{V}$.
Higher entropy corresponds to increased uncertainty in the predicted next-token distribution, while lower entropy corresponds to increased confidence. 
%\paragraph{Log-Likelihood.} 
To quantify the model's uncertainty in the correct next-token, we compute the \emph{token-level log-likelihood} of the ground-truth token $\vy^{*}_t$ as follows:
\begin{align}\label{eq:token-LL}
        C^{*}_{t}= \log P_{\theta}\left(\vy^{*}_t \mid \vy^{*}_{1:t-1}, \mathbf{x} \right) \:.
    \end{align}
%\paragraph{Brier Score.} 
To assess how well-calibrated the model's predicted token distributions are, we adopt the  \emph{Brier Score} \citep{brier1950verification}, and define the token-level calibration as:
\begin{equation}\label{eq:token-calibration}
    \text{Brier}_t = \sum_{k=1}^{\mid\mathcal{V}\mid} \left(P_{\theta}\left(\vy_{k} \mid \vy_{1:t-1}, \mathbf{x}\right) - {\vy_{k}^{*}} \right)^{2} \:,
\end{equation}
which corresponds to the squared Euclidean distance between the predicted token distribution $P_{\theta}\left(\vy \mid \vy_{1:t-1}, \mathbf{x}\right)$ and the one-hot encoded reference token.
\begin{figure}
    \centering
    \vspace{-0.8em}
    \includegraphics[width=0.55\textwidth]{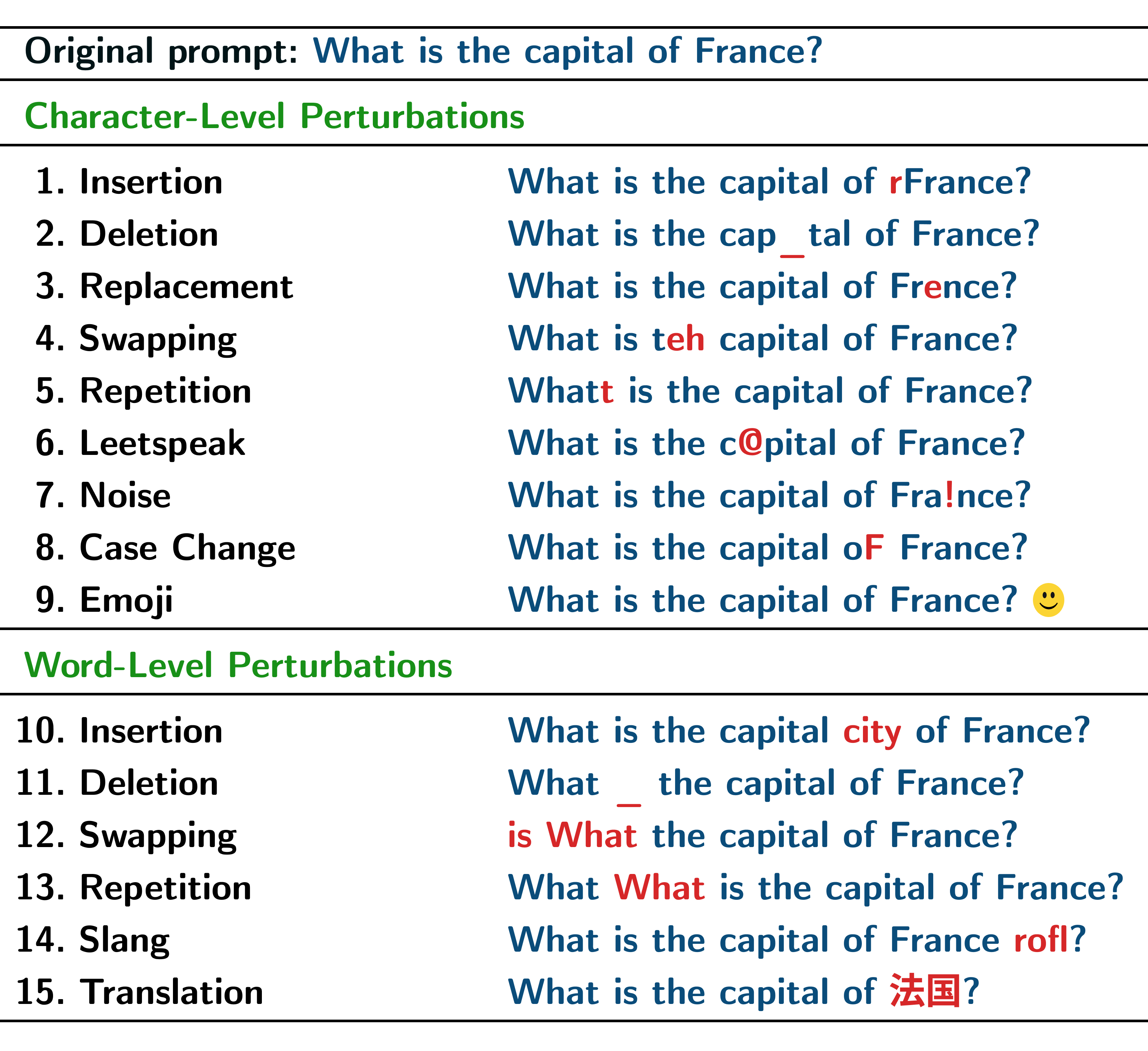}
    \caption{Overview of our perturbations. Illustrated is an example where perturbations with intensity level 1 are applied to a standard question prompt.}
    \label{fig:perturbations_list}
    \vspace{-1em}
\end{figure}
\subsection{Robustness to Input Perturbations}\label{sec:robustness}
In this work, we focus on natural, semantically-preserving perturbations commonly encountered in typed digital communication, such as messaging or chat-based applications.
We introduce character-level and word-level perturbations to the input prompt $\mathbf{x}$. We provide an overview of all input perturbations in \cref{fig:perturbations_list}. We design natural perturbations that are increasingly present in online communication \citep{suhardianto2019colloquail,moradi2021evaluating, hand2022interactions, ackerman2024novel}, but remain underexplored in the robustness evaluation of base and quantized LLMs. We implement the following perturbations: 
\begin{itemize}
    \item \emph{Emoji:} Emojis are increasingly integrated into the written language in interpersonal communication. Current digital communication is more likely to include emojis alongside text rather than traditional emoticons \citep{hand2022interactions}. To evaluate the model's robustness to emojis, we perturb the input prompt by randomly inserting emoji characters. 

    \item \emph{Replacement}: We substitute characters with adjacent keyboard alternatives to simulate realistic typos, using a keyboard layout mapping that maintains case sensitivity \citep{ackerman2024novel}.
    
    \item \emph{Slang:} Slang expressions are commonly used to communicate informally in online dialogues \citep{suhardianto2019colloquail}. To test the model's sensitivity to informal internet language, we introduce perturbations by randomly inserting slang expressions such as "lol", "rofl", and "IMO".  
    
   \item \emph{Insertion:} At the word level, we add contextually relevant words using a language model \citep{liu2019roberta} to predict insertions while preserving text coherence. At the character level, we randomly insert characters.
   
   \item \emph{Leetspeak:} We substitute characters with visually similar numerals or symbols (for example, we replace 'b' with $'6'$). We provide additional examples in \cref{tab:leetspeak}.
   
    %\item \emph{Noise:} We insert punctuations or digits as noise characters distributed randomly in the input prompt.
    \item \emph{Translation:} To simulate the linguistic behavior of individuals with different languages, we randomly translate words to six common languages. Contrary to the semantic-level translation perturbation introduced in \citet{zhu2024promptrobust}, we do not translate phrases back to English.
    
    \item \emph{Deletion:} In the process of deletion, we only remove filler words (e.g., "and", "to", or "actually") that are not crucial for preserving the semantics. For character deletion, we randomly delete characters from the sequence. This perturbation simulates typos that are commonly observed in informal or fast-paced writing \citep{flor2015patterns}.
\end{itemize}
We further design standard character-level and word-level perturbations such as swapping, repetition, and case change \citep{moradi2021evaluating, ackerman2024novel}. 
\citet{ackerman2024novel} test many of the character-level perturbations in adversarial attack scenarios. However, in our work, we focus on a set of non-adversarial perturbations that can
be used to assess the robustness of LLMs.
Additional details on all perturbations are provided in \cref{app:input_perturbations}. 
%To assess the reliability of base and quantized models model under semantically-preserving input perturbations, we evaluate the different reliability metrics using the input prompt $\tilde{\mathbf{x}}$, corresponding to the perturbation of $\mathbf{x}$.
\begin{figure*}[t]
 \vspace{-0.5em}
    \centering
    \includegraphics[width=1.0\textwidth]{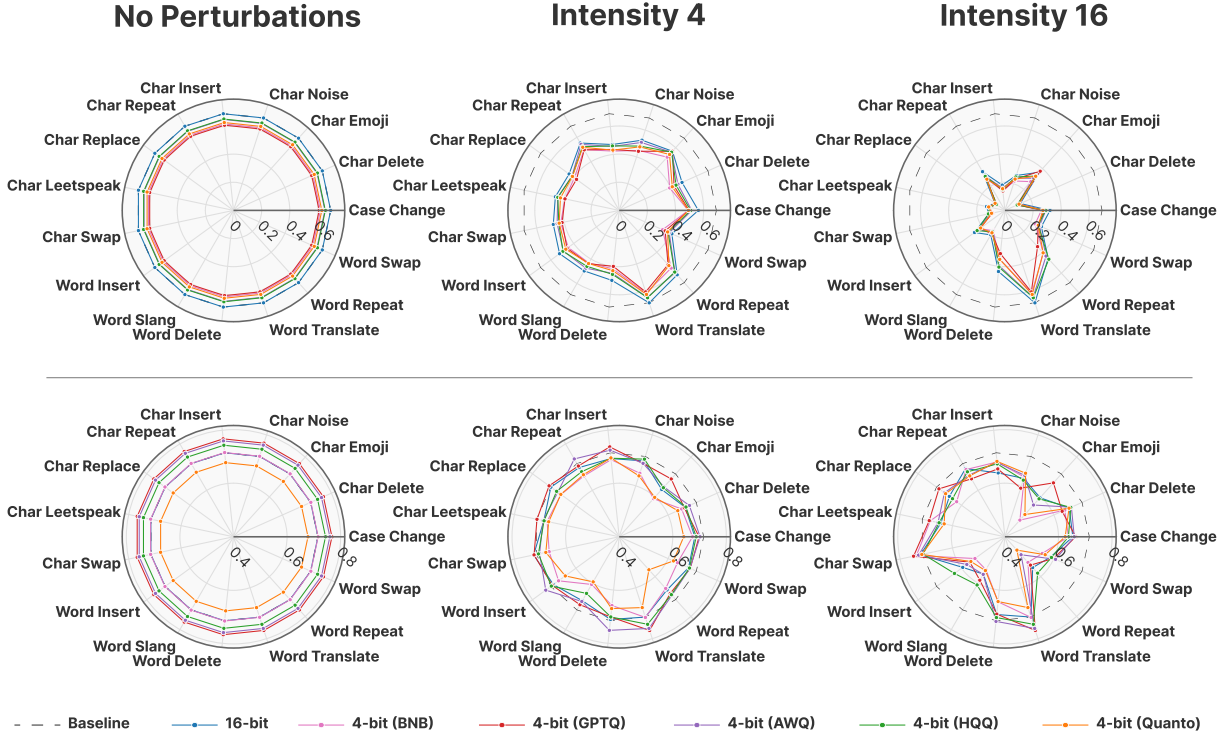}
    \caption{Radar plots of the accuracy (\textbf{Top}) and AUCROC (Entropy) (\textbf{bottom}) across all $15$ character-level and word-level perturbations for two intensities. We evaluate the base LLaMa-3-8B model and five 4-bit quantization methods. Quantized models can provide more reliable uncertainty estimates under natural perturbations compared to their base counterparts, while maintaining a close performance.} 
    \label{fig:radar_triviaqa}
    \vspace{-1em}
\end{figure*}
We provide radar plots of the accuracy and AUCROC (Entropy) on the unperturbed and perturbed TriviaQA dataset across all perturbations in \cref{fig:radar_triviaqa}. We use the full precision LLaMA-3-8B model and five quantized models. We show that 4-bit quantization does not degrade the performance under perturbations. For the reliability evaluation, we report the AUCROC (Entropy) scores, and find that quantization methods, including GPTQ \citep{frantar2022gptq}, AWQ \citep{lin2024awq}, and HQQ \citep{badri_half-quadratic_nodate}, improve reliability under natural perturbations.

\section{{Bit-Level Scaling Trends}}\label{sec:scaling_laws}
Consider two models: \emph{(i)} a model with $K_1$ parameters in $P_1$-bit precision, and \emph{(ii)} a model with $K_2$ parameters in $P_2$-bit precision, such that $K_1 P_1 = K_2 P_2$. Although both models have equal total bit budgets, their reliability characteristics can differ significantly. In this paper, our goal is to characterize how reliability scales with the total number of bits. Using total model bits as a unit for comparison enables a fair model comparison across different parameter counts and bitwidths. We note that we do not propose a formal predictive law in the classical sense; i.e., a closed-form equation that predicts outcomes a priori before evaluation. In prior works \citep{dettmers2023case, frantar2025compression}, scaling laws often refer to empirical regularities observed in model behavior, as a function of model size, compute, or dataset size. These laws are typically fit to the observed data and used to characterize trends. We adopt the same empirical methodology: we empirically characterize consistent trends in reliability behavior as the total number of bits $B$ scales, defined as the product of the number of model parameters and the weight precision (in bits).
Consider a performance metric $\mathcal{L}$ modeled using a \emph{log quadratic scaling law} as follows:
\begin{equation}\label{eq:log-quadratic}
    \mathcal{L}(B) = a (\log(B))^2 + b \log(B) + c\:,
\end{equation}
where $a,b,c \in \mathbb{R}$ are fitted coefficients. This formulation captures nonlinear trends in model performance as the model capacity increases, and generally provides a good fit, as shown in \cref{sec:results}. 

\paragraph{{Why is total model bits a reasonable axis of comparison?}}
{Total model bits provide a unified basis for systematically studying trade-offs between model scale and quantization bitwidth. It is both convenient and easy to measure, while also correlating well with several practical efficiency metrics. In particular, total model bits directly reflect the storage memory and scale linearly with the inference memory. We further discuss in \cref{app:additional_efficiency_metrics} how it relates to inference latency and throughput. \cref{tab:efficiency_metrics} provides a quantitative comparison of efficiency metrics for Qwen models \citep{yang2025qwen3} with comparable total bit budgets.}
\begin{table}[htpb]
\RawFloats
\centering
\caption{We report the total model bits (in GB), the inference memory (in GB), the per-token latency (ms), corresponding to the time spent to generate each token in the output, and the number of decoded tokens per second. We evaluate on a single A100 GPU.}
%\begin{tiny}
\resizebox{0.7\textwidth}{!}{%
\begin{tabular}[S[table-format=2.3]{l|c|c|c|c|c}
\toprule
{Model} & 
{Bitwidth} & 
{\shortstack{Total Model \\ Bits}} & 
{\shortstack{Inference \\ Memory}} & 
{\shortstack{Per-Token \\ Latency}} & 
{\shortstack{Decoded tokens\\per second}} \\
\midrule
{Qwen3-4B} &
{16-bit} &
{8.217}  &
{8.409}  &
{4.103} &
{266.35}
\\
{Qwen3-4B}    &
{8-bit} &
{4.108}  & 
{4.300}  & 
{2.510} & 
{462.66}
\\
{Qwen3-8B}      &
{16-bit} & 
{15.256} & 
{15.476} & 
{9.884} & 
{110.30} \\
{Qwen3-8B}  &
{8-bit}&
{7.628}& 
{7.848}&
{5.806}&
{200.50}
\\
{Qwen3-8B}     &
{4-bit}  & 
{3.814}  &
{4.034}  &
{3.766} &
{339.16} \\
{Qwen3-32B}    &
{4-bit} &
{15.250} &
{15.478} &
{8.719} &
{150.29}
\\
\bottomrule
\end{tabular}

%\end{tiny}
}
\label{tab:efficiency_metrics}
\end{table}

\section{Reliability Assessment of Quantized LLMs}\label{sec:results}
We conduct a comprehensive evaluation of the reliability of quantized LLMs both on unperturbed and perturbed input prompts, across different model scales, families, and precisions. We address the following questions: How does the downstream task performance scale as a function of total model bits? Do the reliability scalings exhibit different trends? What is the optimal precision for trading off reliability and efficiency? Additionally, we conduct ablation studies to investigate whether these scaling behaviors persist across model families and benchmarks.

\subsection{Experimental Setting}\label{sec:exp_setting}
\paragraph{Datasets:}
We consider commonly used benchmarks for evaluating the uncertainty quantification in LLMs across NLG tasks \citep{lin2023generating,kuhn2023semantic}: \textbf{TriviaQA} \citep{joshi2017triviaqa} and \textbf{CoQA} \citep{reddy2019coqa}.
TriviaQA \citep{joshi2017triviaqa} evaluates factual knowledge retrieval with question-answer pairs sourced from trivia competitions. It spans diverse knowledge domains, including history, literature, science, geography, and popular culture. CoQA \citep{reddy2019coqa} (Conversational Question Answering) evaluates contextual understanding through multi-turn conversational exchanges. The dataset consists of conversations from seven diverse domains, including literature, news, and science. Additionally, we evaluate on \textbf{CommonsenseQA} \citep{talmor2018commonsenseqa}, a multiple-choice question-answering dataset designed to assess commonsense knowledge and reasoning capabilities, where the goal is to provide a single answer from five possible answer choices. 
To assess the language modeling capabilities, we evaluate on \textbf{WikiText-2} \citep{merity2016pointer}, \textbf{PTB} \citep{marcus1993building}, and \textbf{C4} \citep{raffel2020exploring}. Additional details are presented in \cref{app:evaluation_datasets}. %We provide additional evaluations on other widely used benchmarks, such as HellaSwag \citep{zellers2019hellaswag}, MMLU \citep{hendrycks2020measuring}, and Arc-Easy \citep{clark2018think}, in \cref{app:performance_scaling}.
{We extend the evaluation to additional benchmarks to assess the emergent abilities of base and quantized models on in-context learning and instruction-following tasks, as well as other NLP tasks such as reasoning and comprehension in \cref{app:additional_performance_scaling}. A summary of all evaluated benchmarks is provided in \cref{tab:benchmarks}.}

\paragraph{Base and quantized LLMs.}
We consider four base pre-trained models \citep{grattafiori2024llama}, including models from the LLaMA-3.2 family: LLaMA-3.2-1B and LLaMA-3.2-3B, and models from the  LLaMA-3 family: LLaMA-3-8B and LLaMA-3-70B. For the quantization, we consider six state-of-the-art quantization methods across various precisions, ranging from 2 bits to 8 bits. These approaches include BitsandBytes \citep{dettmers_2022_bitsandbytes}, AWQ \citep{lin2024awq}, GPTQ \citep{frantar2022gptq}, HQQ \citep{badri_half-quadratic_nodate, badri_towards_nodate}, Quanto \citep{optimumquanto_2023}, and AQLM \citep{egiazarian2024extreme, malinovskii2024pvtuning}. 
For the 2-bit quantization using AQLM, we adopt the AQLM-PV variant, which is a quantized model with additional fine-tuning for improved performance. A complete list of all base models, quantized models, and corresponding precision levels is presented in \cref{app:base_and_quantised_llms}. To better capture performance trends, we fit log-quadratic functions as outlined in \cref{sec:scaling_laws}. 
%\todo{Can remove these two lines}
%{In total, we conduct 5,859 experiments, covering four full-precision models, various quantization methods, bit-level precisions, and $15$ perturbation settings. We report results for the best-performing model in each combination of base model and quantization rate in the scaling plots. For clarity, however, we report results only for the best-performing model in each combination of base model and quantization rate in the scaling plots.} 

\subsection{Bit-Level Scaling Trends}\label{sec:bit-level-scalings-llama}
\begin{figure}[t]
    \centering
    \includegraphics[width=1.0\textwidth]{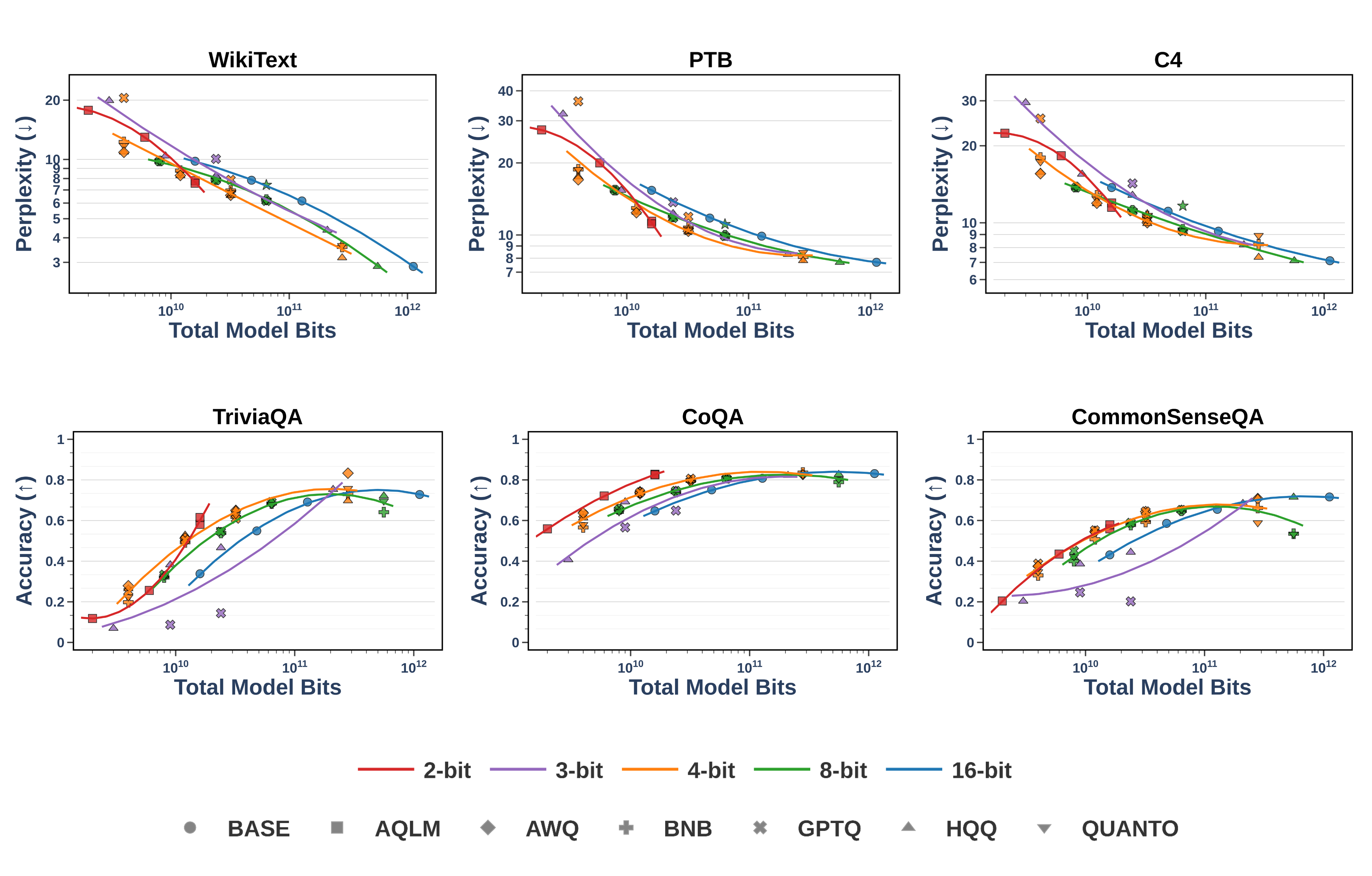}%{figures/performance_scalings_logquadratic.png}
    \caption{Scalings of the perplexity (\textbf{top}) and accuracy (\textbf{bottom}) {for all quantized models and their corresponding full-precision models}. The performance steadily improves with the total number of model bits. 4-bit models offer the best performance-efficiency trade-off given a fixed number of total model bits.} 
    \label{fig:accuracy_and_perplexity}
    %\vspace{-0.5em}
\end{figure}

\paragraph{Downstream task performance scalings.}
We evaluate the downstream performance of base and quantized LLMs on \emph{unperturbed} (i.e., original) datasets.
In \cref{fig:accuracy_and_perplexity}, we present the scaling behavior of the perplexity and the accuracy across different quantization levels and model sizes. First, we observe that the downstream task performance improves with increased total model bits. 
Second, we find that reducing the precision from 16 to 4 bits, for a given total bit budget, consistently improves both the perplexity and the accuracy across all evaluated datasets. 
However, this trend is reversed if we further decrease the precision to 3 bits and 2 bits, aligning with previous observations in \citet{hong2024decoding,dettmers2023case}.
We note that the 2-bit quantization using AQLM \citep{egiazarian2024extreme, malinovskii2024pvtuning} achieves a relatively good performance in the low-bit regime on CoQA and CommonsenseQA. 
For the 3-bit quantization, HQQ \citep{badri_half-quadratic_nodate} achieves the best performance. For the 4-bit setting, both  HQQ \citep{badri_half-quadratic_nodate} and AWQ \citep{lin2024awq} offer the best performance-efficiency trade-off.

\begin{figure}
    \centering
    \includegraphics[width=1.0\textwidth]{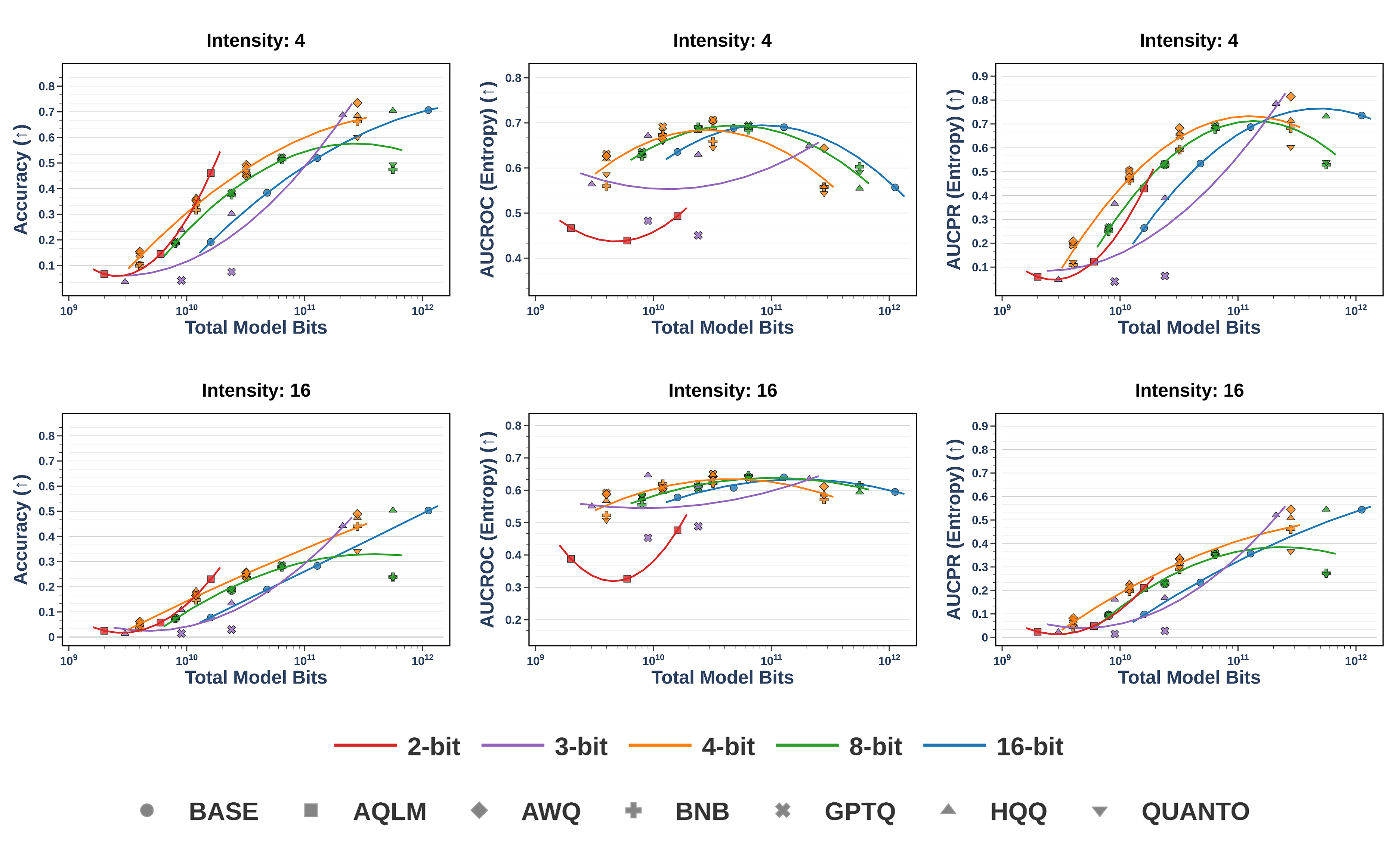}%{figures/All_perturbations_scalings_TriviaQA.png}
    \caption{Bit-level scalings {for all evaluated quantized models and their corresponding full-precision models} on the perturbed TriviaQA dataset, averaged over all 15 perturbations. 
    }\label{fig:all_perturbations_scalings}
    %\vspace{0.5em}
\end{figure}

\paragraph{Reliability scaling trends.}  
%We further provide the scalings on the \emph{perturbed} TriviaQA dataset in \cref{fig:all_perturbations_scalings}. 
To characterize the reliability of base and quantized models, we present the scaling behavior of the accuracy and the reliability metrics on the perturbed TriviaQA dataset in \cref{fig:all_perturbations_scalings}. 
We average the metrics across all 15 perturbations and provide a qualitative comparison for two different perturbation intensities. 
On the one hand, accuracy increases under both unperturbed and perturbed prompts with the total number of model bits. Additionally, 4-bit models offer the strongest robust accuracy {for a fixed bit budget}. We note that the 3-bit HQQ quantization of the LLaMA-3-70B model achieves an accuracy comparable to its 4-bit counterpart. 
On the other hand, bit-level reliability scalings are not linear. More specifically, larger models exhibit lower AUCROC (Entropy), indicating reduced reliability under perturbations.
Extreme 2-bit quantization methods exhibit different scaling trends than moderate precisions. While 4-bit, 8-bit, and 16-bit models have worse uncertainty estimates and calibration in the large-bit regime, the reliability of 2-bit models improves with model scale.  
Additional results are provided in \cref{sec:additional_perturbation_scalings}.
\paragraph{Why does 4-bit quantization offer a favorable {reliability-efficiency trade-off?}}

\begin{wrapfigure}{r}{0.35\textwidth}
    \centering
    \vspace{-1.0em}
    \includegraphics[width=\textwidth]{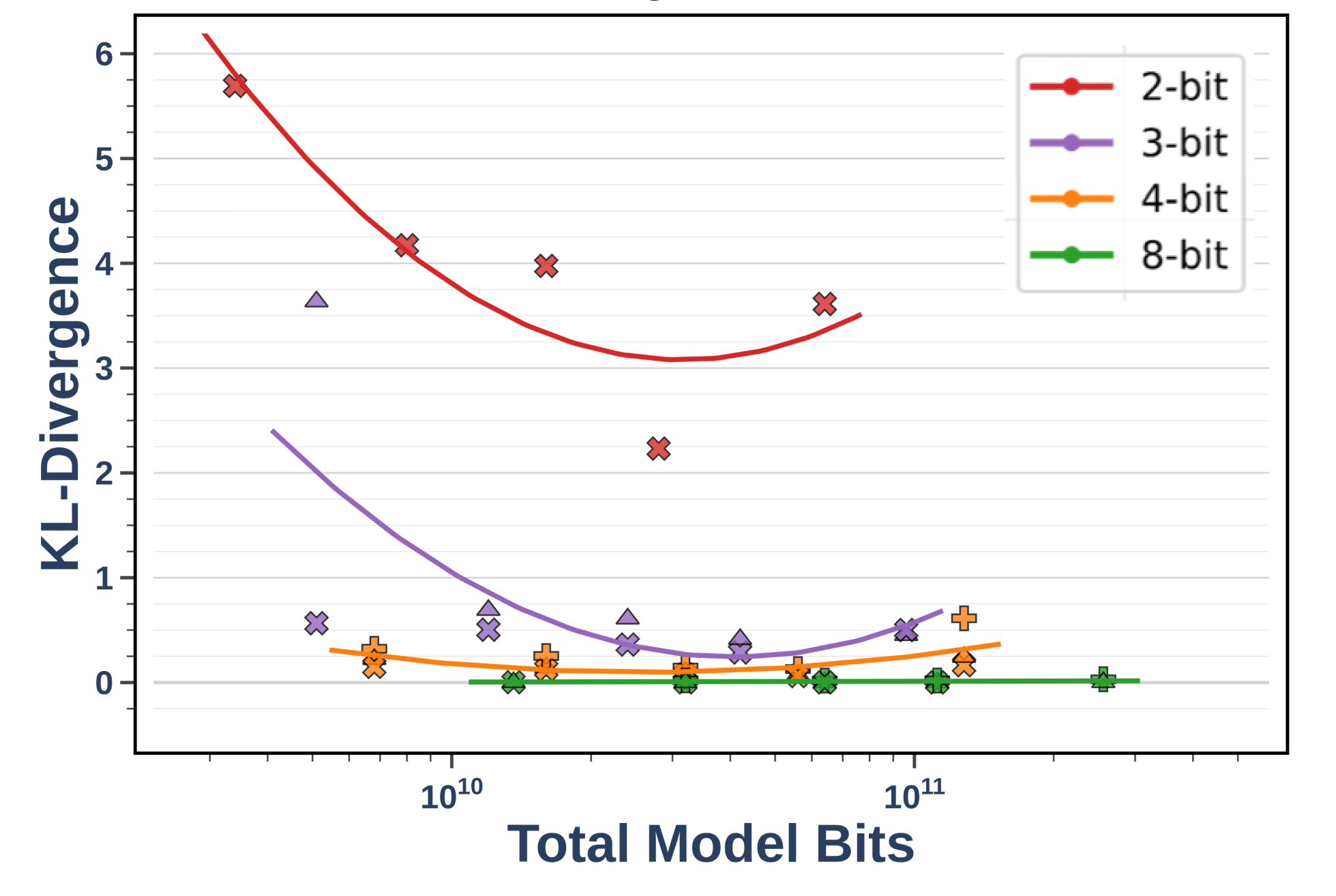}%{figures/KLD_scalings.png}
    \caption{Scaling trends of the KL-Divergence.}
    \label{fig:kld_scaling}
    \vspace{-1.0em}
\end{wrapfigure}

The goal of model quantization is to create a more efficient model from a full-precision base model, while maintaining as close a distance to its full-precision counterpart. While both perplexity and accuracy metrics are essential for evaluating the generalization capabilities of quantized models, they fail to capture the shifts in model behavior that can occur post-compression. To quantify this behavioral shift, we measure the Kullback-Leibler Divergence between the token distributions predicted by the base model, 
$P_{\theta_B}(\vx_t \mid \vx_{<t})$, and those predicted by the compressed model, $P_{{\theta}_C}(\vx_t \mid \vx_{<t})$. 
Let $\mathbf{x} = (\vx_1, \dots, \vx_T) \sim \mathcal{D}_\text{test}$ be a sequence sampled from a test dataset, and $T$ defines the number of tokens in the sequence. The behavioral shift is measured as follows: 
\begin{equation}
\frac{1}{|\mathcal{D}_\text{test}|} \sum_{\mathbf{x} \in \mathcal{D}_\text{test}} \frac{1}{T} \sum_{t=1}^{T} \text{KL}\left(P_{\theta_B}(\vx_t \mid \vx_{<t}) \;\middle\|\; P_{\theta_C}(\vx_t \mid \vx_{<t})\right) \:. 
\end{equation}
In \cref{fig:kld_scaling}, we provide the bit-level scaling trends of the KL-Divergence. We consider 3 LLaMA models with 1B, 3B, and 8B parameters, and quantize them to different bitwidths.
First, we observe that 2-bit quantized models exhibit a substantial behavioral shift relative to their full-precision counterparts, which contributes to their consistently poor performance. For 3-bit quantization, the KL divergence remains noticeably larger than that of 4-bit models. However, as model size increases from 1B to 8B parameters, the behavioral shift induced by 3-bit quantization diminishes. This reduction helps explain the improved downstream performance of larger 3-bit models (see \cref{fig:accuracy_and_perplexity}). In contrast, 4-bit quantization maintains a small KL divergence across model scales. We believe that this small behavioral shift allows 4-bit models to retain strong generalization capabilities while still achieving a substantial reduction in total model bits. 

{While the KL divergence of 8-bit models is better than that of 4-bit models, the key additional factor favoring 4-bit quantized models is their higher model capacity within a fixed total-bit budget. A 4-bit model matched to an 8-bit model in total bits has roughly twice the number of parameters. Our results suggest that, in this regime, the benefit of increased model capacity can outweigh the benefit of higher weight precision: 4-bit models remain sufficiently close to their full-precision counterparts while benefiting from a larger parameter count. Below 4 bits, this trade-off changes: the behavioral shift induced by aggressive quantization becomes too large to be compensated by additional parameters.}

\begin{figure}[]
    \centering
    \begin{subfigure}{\linewidth}
        \centering
        \includegraphics[width=\linewidth]{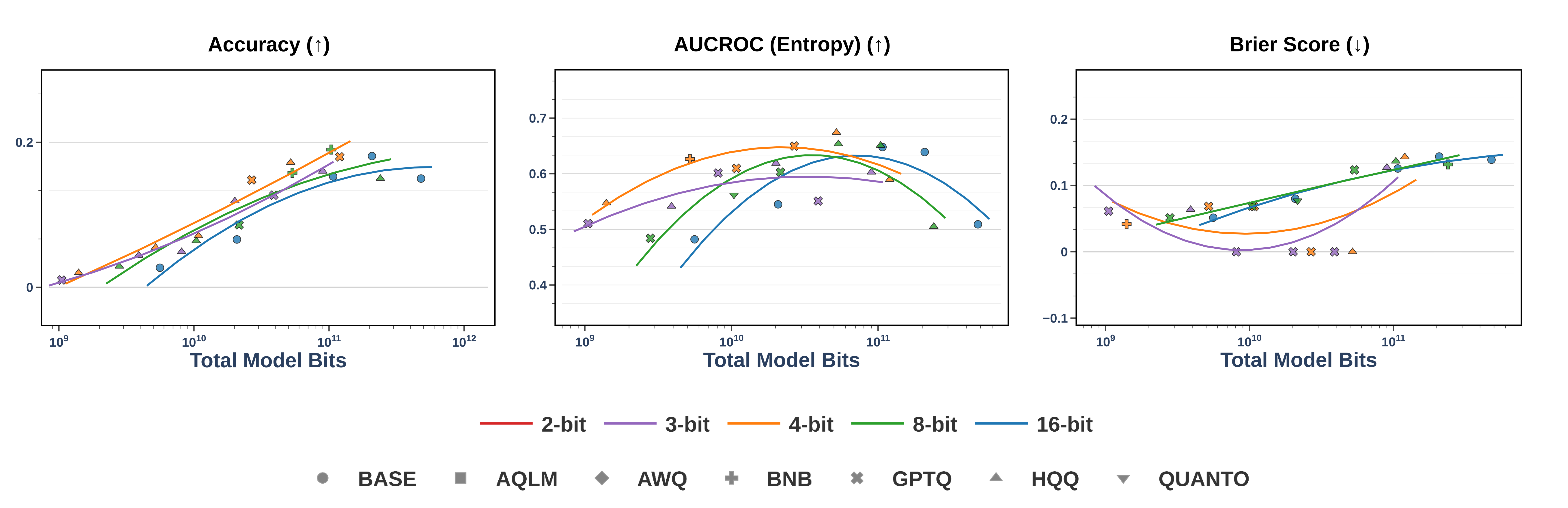}
        \caption{Scaling trends of OPT models on TriviaQA.}
        \label{fig:opt_wordslang}
    \end{subfigure}
    \vspace{0.1em}
    \begin{subfigure}{\linewidth}
        \centering
        \includegraphics[width=\linewidth]{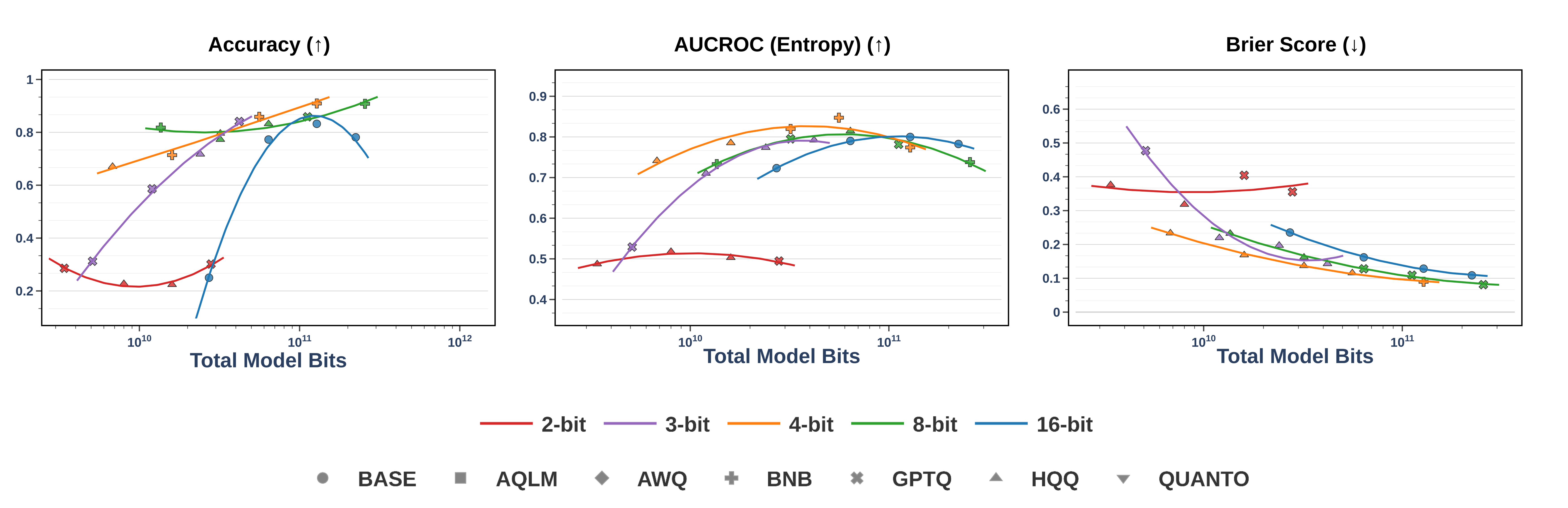}
        \caption{Scaling trends of Qwen3 models on CEval.}
        \label{fig:qwen3_ceval}
    \end{subfigure}
    \caption{Scaling trends of OPT and Qwen3 models across different benchmarks.}
    \label{fig:opt_qwen_scaling}
    \vspace{-1em}
\end{figure}
\paragraph{{Are the scaling trends consistent across diverse model architectures?}} 
%{In addition to models from the LLaMA-3 series, we consider models from OPT and the new Qwen3 model families. For OPT, we consider six base models, with parameters in {350M, 1.3B, 2.7B, 13B, 30B}, and provide qualitative results in \cref{fig:opt_wordslang}. For Qwen3, we consider models with sizes in {4B, 8B, 14B, 32B}, and provide results in \cref{fig:qwen3_ceval}.}
%{The results indicate that the 4-bit sweet-spot is generally consistent across diverse model backbones. We also find that the performance and reliability scaling trends are consistent with previous observations. In particular, the accuracy scales linearly with total model bits, with 4-bit models outperforming all other precisions, including the full-precision models, for a specific budget of total model bits. For the reliability scalings, we observe non-linear trends, particularly for AUCROC (Entropy) and AUCPR(Entropy), and the Brier scores.}
%{For OPT models,} both GPTQ and HQQ yield the best performance among 4-bit quantization, with GPTQ yielding the best performance at 3-bit.
%{For Qwen models, ..}
%{Additional evaluations of Qwen models on different widely used benchmarks are presented in \cref{app:additional_results_qwen}. }
{We extend our analysis beyond the LLaMA-3 family to include models from OPT \citep{zhang2022opt} and Qwen3 \citep{yang2025qwen3} series. For OPT,} we evaluate six base models with parameter counts of 350M, 1.3B, 2.7B, 13B, and 30B, and present qualitative results in \cref{fig:opt_wordslang}. {For Qwen3, we consider models of sizes 4B, 8B, 14B, and 32B, and present results in \cref{fig:qwen3_ceval}. Across different architectures, we observe a consistent 4-bit sweet spot: 4-bit quantization generally achieves the best trade-off between accuracy, reliability, and efficiency. Moreover, the scaling behavior aligns with our earlier findings. 
%The accuracy scales linearly with total model bits, and under a fixed bit-budget, 4-bit models consistently outperform higher-precision models. In contrast, reliability metrics exhibit non-linear trends. 
Interestingly, for Qwen models, we find that the extreme 3-bit quantization of moderate-to-large models yields favorable performance.} 
For OPT models, HQQ and GPTQ exhibit the best performance among 4-bit quantization methods, with GPTQ achieving the best results among 3-bit quantizations. {For Qwen models, HQQ and BitsandBytes outperform other PTQ approaches in the 4-bit setting, whereas GPTQ and HQQ provide the best performance under 3-bit compression. 
Additional evaluations on widely used benchmarks for Qwen models are presented in \cref{app:additional_performance_scaling}.}
\begin{figure}[]
    \centering

    \begin{subfigure}{\textwidth}
        \centering
        \includegraphics[width=\textwidth]{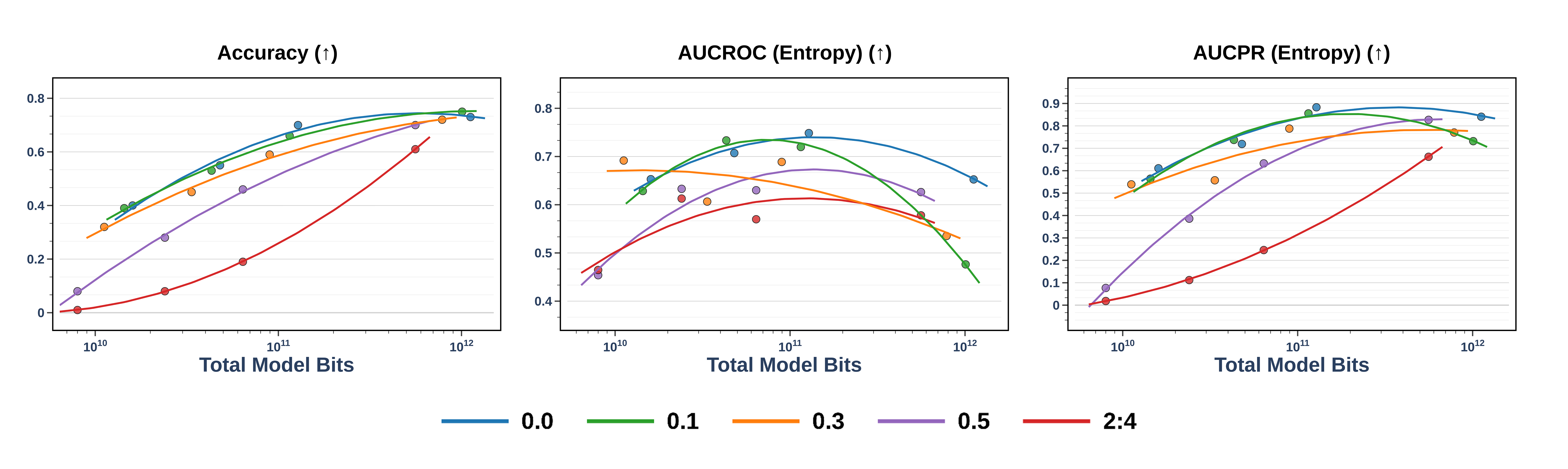}
        \vspace{-0.2em}
        \caption{Scaling trends of models pruned using SparseGPT to different sparsities.}
        \label{fig:sparsegpt}
    \end{subfigure}
    \vspace{0.5em}
    \begin{subfigure}{\textwidth}
        \centering
        \includegraphics[width=\textwidth]{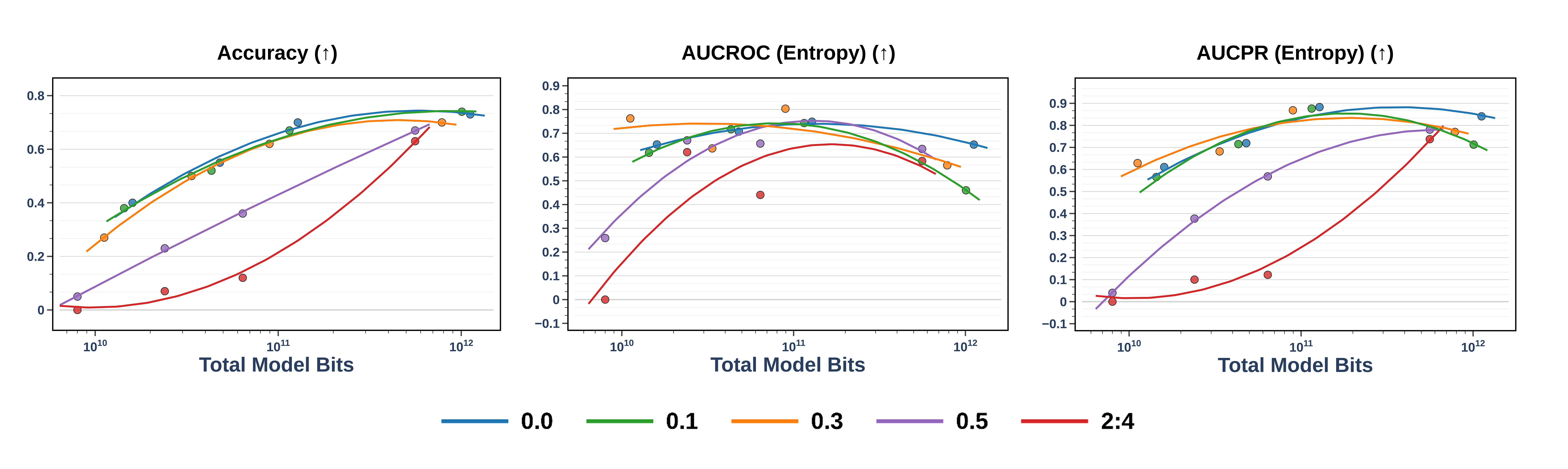}
        \vspace{-0.2em}
        \caption{Scaling trends of models pruned using Wanda to different sparsities.}
        \label{fig:wanda}
    \end{subfigure}
    \caption{Scaling trends of pruned LLaMA models using SparseGPT and Wanda.}
    \label{fig:pruning_scaling}
    \vspace{-1em}
\end{figure}
\paragraph{{How do pruning strategies affect the reliability scaling trends of LLMs?}}
{While our primary focus is to characterize the scaling behavior of PTQ techniques across various model sizes and bitwidths, motivated by their methodological diversity and growing adoption, we also examine pruning, another widely used post-training compression strategy.}
{In particular, we evaluate two state-of-the-art pruning techniques for LLMs: Wanda \citep{sun2023simple} and SparseGPT \citep{frantar2023sparsegpt}.} 
{We prune four base LLaMA models at sparsity levels of {0.1, 0.3, 0.5}, and additionally apply semi-structured 2:4 pruning. The N:M sparsity pattern enforces that only N weights remain non-zero within each block of M consecutive weights. Unlike unstructured pruning, N:M pruning can provide actual hardware acceleration.} 
{The corresponding scaling trends for SparseGPT and Wanda are presented in \cref{fig:sparsegpt} and \cref{fig:wanda}, respectively, where base models have zero sparsity. We find that as the sparsity increases, both performance and reliability degrade.
While unstructured pruning at low sparsity levels can preserve a performance close to that of their base counterparts, it does not yield substantial gains as observed with 4-bit quantization. This is consistent with prior findings \citep{harma2024effective}, which suggest that quantization-induced error is lower than that introduced by sparsification; and therefore, quantized LLMs generally outperform pruned LLMs. While both 50\% sparse and semi-structured models exhibit noticeable degradation in reliability, we find that their performance improves significantly as model scale increases. A similar trend is observed for 3-bit quantized models, as shown in \cref{fig:all_perturbations_scalings}.}

\subsection{Discussion and limitations}
In this paper, we provide the first comprehensive reliability evaluation of base and quantized models by studying the underlying bit-level scaling trends across different model scales and bitwidths. 
While prior work \citet{jaiswal2023compressing,dutta2024accuracy,dettmers2023case} focus exclusively on zero-shot performance of quantized models, we argue that the downstream task performance is insufficient. Notably, although the performance scales linearly with the total number of bits, with 4-bit quantization generally yielding the highest accuracy, the reliability trends are not linear: A peak occurs for 4-bit models, suggesting that a favorable reliability-efficiency trade-off can be achieved without resorting to the largest model or the highest precision. We further examine the robustness of base and quantized models under natural input perturbations and reveal that 4-bit quantization can enhance the robustness to semantically preserving perturbations that occur in typed digital communication. Our findings are consistent across diverse model architectures and datasets. We further examine the scalings of two pruning methods \citep{sun2023simple, frantar2023sparsegpt}, and find that sparsification does not offer significant reliability gains as observed with quantization, aligning with recent findings \citep{harma2024effective}. 

For future work, exploring how model calibration, multi-shot prompting, and model fine-tuning can affect the reliability scalings of quantized LLMs is crucial. Additionally, an interesting direction is studying which hyperparameters improve the reliability scalings for a specific number of model parameters and bit precision. We focus on post-training quantization approaches as they are widely adopted. Future work can extend this study to other compression approaches, such as Quantization-Aware Training (QAT) techniques, which may reveal different trends. This is particularly interesting given the potential of QAT to improve the performance of extreme quantization \citep{ma2024era}, see \cref{sec:QAT}.

\section{Conclusion}\label{sec:conclusion}
We present a comprehensive evaluation of quantized LLMs across key reliability dimensions, focusing specifically on uncertainty and robustness to semantically-preserving input perturbations. While prior studies have primarily emphasized evaluating the accuracy of quantized models on standard benchmarks, our findings reveal that this narrow focus overlooks critical safety considerations. By studying reliability scaling trends, we show that reliability does not necessarily scale monotonically with the total number of model bits. An optimal reliability-efficiency trade-off can be achieved without resorting to the highest precision or quantizing the largest base model. We emphasize the potential of quantization to improve model reliability, making it a promising approach for deploying trustworthy LLMs in practical settings.

\subsubsection*{Broader Impact Statement}
Our research examines the reliability of quantized large language models across multiple dimensions to support their safe and trustworthy deployment in real-world applications. While quantization methods help reduce the energy and emissions impact of machine learning applications, which is an urgent environmental challenge, they may introduce new challenges, such as decreased robustness to semantically-preserving input perturbations. We highlight the need for more comprehensive evaluations to ensure that quantized models are robust and reliable, thereby meeting the standards required for responsible deployment.

\subsubsection*{Acknowledgments}
SA, SG, and BC acknowledge funding from the German Federal Ministry of Research, Technology and Space (BMFTR) under grant agreement No. 01IS24072C (COMFORT). This work was also supported by the DAAD programme Konrad Zuse Schools of Excellence in Artificial Intelligence, sponsored by the Federal Ministry of Education and Research.
%\newpage
\bibliography{main}
\bibliographystyle{tmlr}

\appendix
%\appendix
%\onecolumn
\newpage
\section{Additional details on the experimental setting}
\subsection{Quantization methods}\label{app:base_and_quantised_llms}
%\subsection{Additional details on the experimental setting}
We present the complete list of base models and quantized models used in our evaluations in \cref{tab:quantized_models_used}. We consider different quantization methods, including BNB \citep{dettmers_2022_bitsandbytes}, AWQ \citep{lin2024awq}, GPTQ \citep{frantar2022gptq}, HQQ \citep{badri_half-quadratic_nodate, badri_towards_nodate}, Quanto \citep{optimumquanto_2023}, AQLM \citep{egiazarian2024extreme, malinovskii2024pvtuning}, QuaRot \citep{ashkboos_quarot_2024},  and QoQ \citep{lin_qserve_2024}, applied across 4 LLaMA models from the LLaMA-3 family. Not all bitwidth configurations are adopted for every base model due to compatibility constraints and computational limitations. In total, our evaluation encompasses 63 distinct model configurations that we evaluate in the reliability assessment framework. 
As we only report the best-performing model for each bit width in \cref{sec:results}, both QuaRot and Quanto are not included in the scaling laws figures as they do not outperform other quantization techniques. In addition, we note that we use AQLM-PV for the 2-bit quantization, which corresponds to AQLM quantization with an additional PV tuning step. AQLM-PV provides increased performance improvement compared to AQLM. We run $63 * 3 * (1 + 15 * 2)= 5859$ experiments in total, where 63 is the number of models, 3 is the number of the QA datasets, 15 is the number of evaluations, and 2 corresponds to the two perturbation intensities 4 and 16.

For the quantization using Bitsandbytes \citep{dettmers_2022_bitsandbytes}, HQQ \citep{badri_half-quadratic_nodate}, Quanto \citep{optimumquanto_2023} and GPTQ (3-bit), we use the Hugging Face Transformers \citep{wolf2019huggingface} default implementation, licensed under the Apache-2.0 license. GPTQ is licensed under the Apache-2.0 license, while AWQ is licensed under the MIT license. All AQLM, QuaRot, and QoQ models are loaded directly from HuggingFace, licensed under the Apache License 2.0. The LLaMA-3 and LLaMA-3.2 model families we used in our experiments are licensed under the Llama 3 Community License Agreement.
\begin{table}[h]
\centering
\caption{Quantization techniques used in our reliability evaluation framework. \textcolor{black!80!black}{\checkmark} corresponds to models included in the reliability assessment, and \textcolor{black}{\texttimes} corresponds to quantization-precision combinations we do not include.}
\label{tab:quantized_models_used}
\begin{small}
% \resizebox{.5\textwidth}{!}{
\begin{tabular}{llcccc}
\toprule
\multicolumn{2}{c}{{Quantization Method}} &
\multicolumn{4}{c}{{Model Size}} \\
\cmidrule(lr){1-2} \cmidrule(lr){3-6}
{Method} & {Bit} & {1B} & {3B} & {8B} & {70B} \\
\midrule
Base & 16-bit & \textcolor{black!80!black}{\checkmark} & \textcolor{black!80!black}{\checkmark} & \textcolor{black!80!black}{\checkmark} & \textcolor{black!80!black}{\checkmark} \\
\midrule
AQLM & 2-bit & \textcolor{black}{\texttimes} & \textcolor{black}{\texttimes} & \textcolor{black!80!black}{\checkmark} & \textcolor{black}{\texttimes} \\
\midrule
AQLM-PV & 2-bit & \textcolor{black!80!black}{\checkmark} & \textcolor{black!80!black}{\checkmark} & \textcolor{black!80!black}{\checkmark} & \textcolor{black}{\texttimes} \\
\midrule
AWQ & 4-bit & \textcolor{black!80!black}{\checkmark} & \textcolor{black!80!black}{\checkmark} & \textcolor{black!80!black}{\checkmark} & \textcolor{black!80!black}{\checkmark} \\
\midrule
\multirow{2}{*}{BNB} & 8-bit & \textcolor{black!80!black}{\checkmark} & \textcolor{black!80!black}{\checkmark} & \textcolor{black!80!black}{\checkmark} & \textcolor{black!80!black}{\checkmark} \\
 & 4-bit & \textcolor{black!80!black}{\checkmark} & \textcolor{black!80!black}{\checkmark} & \textcolor{black!80!black}{\checkmark} & \textcolor{black!80!black}{\checkmark} \\
\midrule
\multirow{4}{*}{GPTQ} & 8-bit & \textcolor{black!80!black}{\checkmark} & \textcolor{black!80!black}{\checkmark} & \textcolor{black!80!black}{\checkmark} & \textcolor{black}{\texttimes} \\
 & 4-bit & \textcolor{black!80!black}{\checkmark} & \textcolor{black!80!black}{\checkmark} & \textcolor{black!80!black}{\checkmark} & \textcolor{black}{\texttimes} \\
 & 3-bit & \textcolor{black!80!black}{\checkmark} & \textcolor{black!80!black}{\checkmark} & \textcolor{black!80!black}{\checkmark} & \textcolor{black}{\texttimes} \\
& 2-bit & \textcolor{black!80!black}{\checkmark} & \textcolor{black!80!black}{\checkmark} & \textcolor{black!80!black}{\checkmark} & \textcolor{black}{\texttimes} \\
\midrule
\multirow{4}{*}{HQQ} & 8-bit & \textcolor{black!80!black}{\checkmark} & \textcolor{black!80!black}{\checkmark} & \textcolor{black!80!black}{\checkmark} & \textcolor{black!80!black}{\checkmark} \\
 & 4-bit & \textcolor{black!80!black}{\checkmark} & \textcolor{black!80!black}{\checkmark} & \textcolor{black!80!black}{\checkmark} & \textcolor{black!80!black}{\checkmark} \\
 & 3-bit & \textcolor{black!80!black}{\checkmark} & \textcolor{black!80!black}{\checkmark} & \textcolor{black!80!black}{\checkmark} & \textcolor{black!80!black}{\checkmark} \\
 & 2-bit & \textcolor{black!80!black}{\checkmark} & \textcolor{black!80!black}{\checkmark} & \textcolor{black!80!black}{\checkmark} & \textcolor{black!80!black}{\checkmark} \\
\midrule
QoQ & 4-bit & \textcolor{black}{\texttimes} & \textcolor{black}{\texttimes} & \textcolor{black!80!black}{\checkmark} & \textcolor{black}{\texttimes} \\
\midrule
\multirow{3}{*}{Quanto} & 8-bit & \textcolor{black!80!black}{\checkmark} & \textcolor{black!80!black}{\checkmark} & \textcolor{black!80!black}{\checkmark} & \textcolor{black!80!black}{\checkmark} \\
 & 4-bit & \textcolor{black!80!black}{\checkmark} & \textcolor{black!80!black}{\checkmark} & \textcolor{black!80!black}{\checkmark} & \textcolor{black!80!black}{\checkmark} \\
 & 2-bit & \textcolor{black!80!black}{\checkmark} & \textcolor{black!80!black}{\checkmark} & \textcolor{black!80!black}{\checkmark} & \textcolor{black!80!black}{\checkmark} \\
\midrule
\multirow{2}{*}{QuaRot} & 8-bit & \textcolor{black}{\texttimes} & \textcolor{black}{\texttimes} & \textcolor{black!80!black}{\checkmark} & \textcolor{black}{\texttimes} \\
 & 4-bit & \textcolor{black}{\texttimes} & \textcolor{black}{\texttimes} & \textcolor{black!80!black}{\checkmark} & \textcolor{black}{\texttimes} \\
\bottomrule
\end{tabular}
% }
\end{small}
\end{table}

\newpage
\subsection{Evaluation datasets and generation}\label{app:evaluation_datasets}
\begin{table}[]
\centering
\caption{Summary of the evaluated benchmarks. We provide the corresponding bit-level inference scalings in \cref{sec:results} and \cref{app:additional_performance_scaling}.}
\begin{small}
\begin{tabular}{ll}
\toprule
\textbf{Task \& Ability} & \textbf{Benchmark} \\ 
\midrule

\multirow{3}{*}{\textbf{Language Modeling}}
    & PTB \citep{marcus1993building} \\
    & C4 \citep{raffel2020exploring} \\
    & WikiText2 \citep{merity2016pointer} \\
\midrule

\multirow{2}{*}{\textbf{\shortstack[l]{Open-Ended QA Generation for \\ General Knowledge \& Dialog}}}
    & TriviaQA \citep{joshi2017triviaqa} \\
    & CoQA \citep{reddy2019coqa} \\
\midrule

\multirow{2}{*}{{\textbf{In-Context Learning}}}
    & MMLU \citep{hendrycks2020measuring} \\
    & CEval \citep{huang2023c} \\
\midrule

\multirow{2}{*}{\textbf{Instruction Following}}
    & HellaSwag \citep{zellers2019hellaswag} \\
    & ARC \citep{clark2018think} \\
\midrule

\multirow{2}{*}{\textbf{Reasoning}}
    %& LogiQA \\
    & PIQA \citep{bisk2020piqa} \\
    & CommonsenseQA \citep{talmor2018commonsenseqa} \\
\midrule

\multirow{1}{*}{\textbf{Understanding}}
    %& LAMBADA \\
    & RACE \citep{lai2017race} \\
\bottomrule

\end{tabular}
\end{small}
\label{tab:benchmarks}
\end{table}
\begin{figure}[]
 %\vspace{-1em}
    \centering    
    \includegraphics[width=0.9\textwidth]{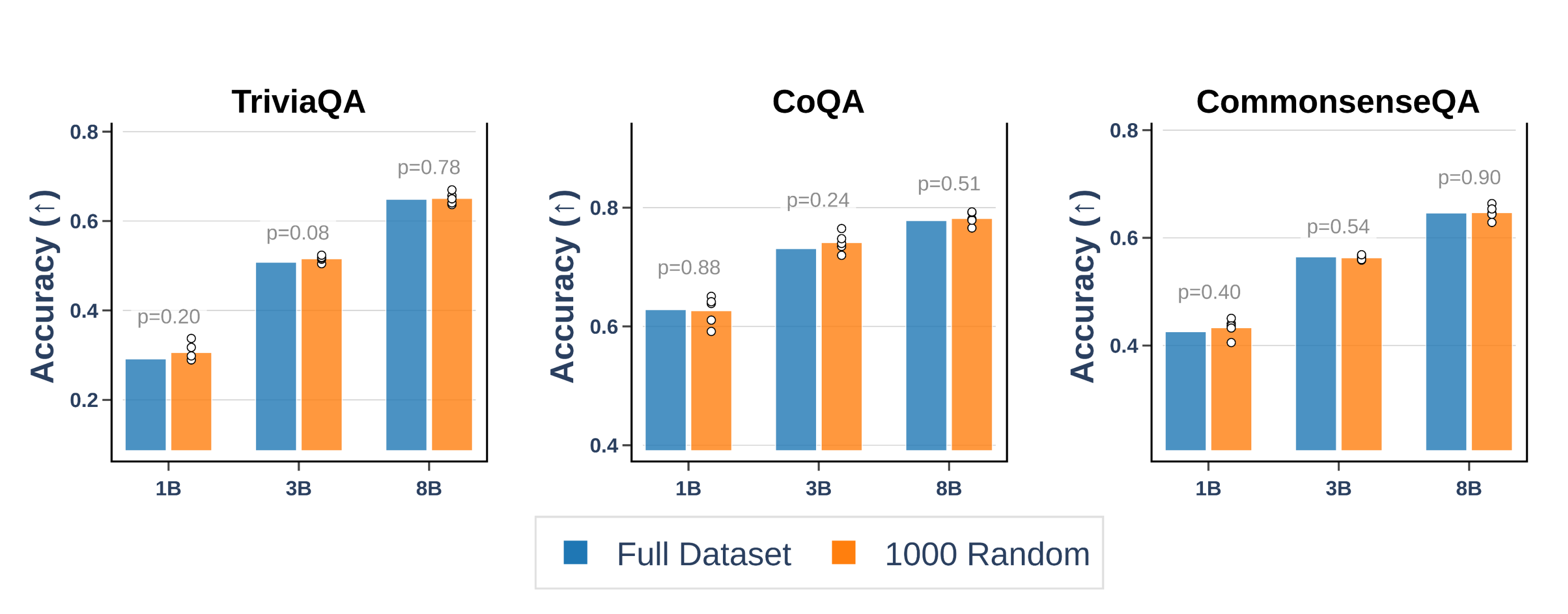}
    \caption{Comparison of the accuracy when evaluating on the full datasets versus evaluating on 1000 randomly sampled prompts (averaged over 5 random seeds).} 
    \label{fig:full_datasets_vs_1000}
    %\vspace{-1em}
\end{figure}
For the evaluation, we use two widely-used datasets for assessing the reliability of LLMs, TriviaQA \citep{joshi2017triviaqa} and CoQA \citep{reddy2019coqa}. TriviaQA consists of 95,000 question-answer pairs to assess reading comprehension capabilities, while CoQA consists of 8,000 question-answer pairs. We further evaluate on the multiple-choice question answering CommonsenseQA \citep{lin2004automatic} dataset. We also evaluate on additional widely used benchmarks such as  HellaSwag \citep{zellers2019hellaswag}, MMLU \citep{hendrycks2020measuring}, and Arc-Easy \citep{clark2018think}. {We summarize all evaluation benchmarks in \cref{tab:benchmarks}}. For all datasets, we limit the evaluation to 1000 uniformly sampled prompts, due to the increased number of evaluation combinations as we consider four base models, eight quantization models, and five different bit precision levels. 
We randomly sample 1000 samples from the datasets for 5 random seeds, and evaluate the accuracy. We compare the accuracies over the sampled sets with the accuracy averaged across the entire datasets. We show in \cref{fig:full_datasets_vs_1000} that the accuracy on the subsets is close to the full dataset evaluation.
During generation, we limit the generated sequence length to 20 tokens per input prompt. During the generation, we use multinomial sampling with a temperature parameter of 0.7.

We conduct our experiments using GTX 1080 Ti (11GB) and A100 (40GB) GPUs. For smaller models, specifically all 1B and 3B models (excluding AQLM and GPTQ variants), we use a single GTX 1080 Ti GPU. 
Medium-sized models, including the 1B and 3B AQLM variants, GPTQ, the 8-bit versions, and all 8B models, are evaluated on a single A100 GPU. For the largest 70B models, we use five A100 GPUs in parallel to handle loading and evaluation efficiently. For the evaluation runtime given a single perturbation (or no perturbation), the average evaluation time is approximately 5 minutes on the GTX 1080 Ti for the small models, and 4 minutes on the A100, and roughly 24 minutes for the 70B models on the five A100 GPUs.

\section{Additional details on the perturbations}\label{app:input_perturbations}
In addition to the character-level and word-level perturbations described in \cref{sec:robustness}, we provide additional details on additional perturbations in the following:
\begin{table}[]
\centering
\caption{Examples of leetspeak character perturbations.}
\label{tab:leetspeak}
\begin{small}
\begin{tabular}{c|c|c|c}
    \toprule
    \multicolumn{4}{c}{\textbf{Leetspeak Character Substitutions}} \\
    \midrule
    A $\,\rightarrow\,$ @ & 
    a $\,\rightarrow\,$ @ & 
    B $\,\rightarrow\,$ 8 & 
    b $\,\rightarrow\,$ 6 \\[1.2ex]
    
    C $\,\rightarrow\,$ ( & 
    c $\,\rightarrow\,$ © & 
    D $\,\rightarrow\,$ ) & 
    d $\,\rightarrow\,$ $\delta$ \\[1.2ex]
    
    E $\,\rightarrow\,$ 3 & 
    e $\,\rightarrow\,$ € & 
    G $\,\rightarrow\,$ 6 & 
    g $\,\rightarrow\,$ 9 \\[1.2ex]
    
    H $\,\rightarrow\,$ \# & 
    I $\,\rightarrow\,$ 1 & 
    i $\,\rightarrow\,$ ! & 
    J $\,\rightarrow\,$ 7 \\[1.2ex]
    
    K $\,\rightarrow\,$ $|<$ & 
    L $\,\rightarrow\,$ £ & 
    l $\,\rightarrow\,$ $|$ & 
    O $\,\rightarrow\,$ 0 \\[1.2ex]
    
    o $\,\rightarrow\,$ ° & 
    P $\,\rightarrow\,$ ? & 
    R $\,\rightarrow\,$ ® & 
    S $\,\rightarrow\,$ 5 \\[1.2ex]

    s $\,\rightarrow\,$ § & 
    T $\,\rightarrow\,$ 7 & 
    w $\,\rightarrow\,$ $\omega$ & 
    X $\,\rightarrow\,$ $\times$ \\
    \bottomrule
\end{tabular}
\end{small}
\end{table}
\begin{itemize}
    \item Deletion: For character-level perturbation, we randomly remove characters from the sentence; at the word level, we remove words from text, primarily targeting filler words (e.g., "and", "to", "also", "actually") to preserve the important part of the prompt \citep{moradi2021evaluating}. We use 508 filler words in total. 
    \item Leetspeak: Substitutes characters with visually similar numerals or symbols (like replacing 'e' with '3'). We provide additional examples in \cref{tab:leetspeak}. We use 93 Leetspeak examples.
    \item Noise: Inserts punctuation or digits as noise characters distributed randomly throughout the text, using a configurable noise character set.
     \item Replacement: Substitutes characters with adjacent keyboard alternatives to simulate realistic typos, using a keyboard layout mapping maintaining case sensitivity \citep{ackerman2024novel}.
    \item Swapping: Exchanges adjacent characters to simulate typographical errors; at the word level, exchanges adjacent words while preserving punctuation and maintaining the question-answer prompt structure \citep{moradi2021evaluating, wang2023arelarge}.
    \item Repetition: Duplicates characters at random positions; at the word level, duplicates words while handling punctuation preservation and maintaining overall text structure \citep{moradi2021evaluating}.
    \item Case Change: Alters letter case patterns, toggling between uppercase and lowercase only for alphabetic characters \citep{moradi2021evaluating}.
    \item Emoji: A novel perturbation type that inserts emoji characters within text, distributing them across words using a configurable emoji set. We use 115 emoji characters in total.
    \item Slang: Inserts internet slang expressions (like "lol", "rofl", "imo") using a predefined dictionary of common terms, distributing insertions randomly throughout the text.
    \item Translation: Translates words and phrases through Google Translate API across seven languages (Spanish, French, German, Italian, Russian, Chinese, Japanese). As opposed to the semantic-level translation perturbation introduced in \citet{zhu2024promptrobust}, our method does not translate phrases back to English, simulating multi-lingual user outputs.
    \item Insertion: Randomly inserts characters between existing characters at the character level; at the word level, adds contextually relevant words using RoBERTa masked language model to predict insertions while preserving text coherence. This approach is methodologically similar to the synonym insertion implemented by \citet{moradi2021evaluating}, but only adds words and does not change them.
\end{itemize}
For the leetspeak character perturbations, we provide examples in \cref{tab:leetspeak}.
Each perturbation is applied at varying intensity levels of 4 and 16 to systematically evaluate model robustness across multiple intensities. We note that for the word perturbations presented in \cref{fig:perturbations_list}, the number of perturbed words in the input prompt corresponds to the minimum between the perturbation intensity and the length of the input prompt.

\section{{Additional Efficiency Metrics}}\label{app:additional_efficiency_metrics}

{In this section, we examine how the total model bits metric relates to several commonly used efficiency measures, including inference memory, latency, and throughput. A quantitative comparison of these metrics is provided in \cref{tab:efficiency_metrics} for different models from the Qwen3 model family. As introduced in \cref{sec:scaling_laws}, the total number of bits 
$B$ is defined as the product of the number of model parameters and their weight precision (in bits). This quantity corresponds directly to the model’s storage memory. The total model bits further maps to the inference memory, measured as:
\begin{equation}
    \text{Inference memory} = P \times \text{Avg-Bitwidth} + \text{Overhead},
\end{equation}
where $P$ is the total number of parameters, and the Overhead term accounts for the additional memory used during inference (e.g., KV cache, activation). In practice, the overhead is generally negligible relative to the storage memory, which implies that inference memory scales almost linearly with the total model bits.} 

{
For inference latency, the relationship with total model bits is less direct but still significant. The overall latency is determined by two phases: (1) the pre-fill phase and (2) the decode phase. 
During prefill, the model processes the input sequence and constructs the key–value (KV) cache. This stage is typically compute-intensive and can fully utilize GPU compute units.
During the decode phase, the model generates output tokens sequentially. Each token is predicted based on the previously generated tokens and the information stored in the KV cache from the prefill stage. 
Lower-precision numbers can improve speed because they reduce the amount of data that needs to be moved in memory and allow computations to be executed more quickly. For instance, on modern GPUs such as H100, 8-bit operations can achieve significantly higher FLOPS than 32-bit operations. However, this improvement also depends on specialized hardware support \footnote{\url{https://resources.nvidia.com/en-us-gpu-resources/h100-datasheet-24306}}. For future work, it would be interesting to investigate scaling laws as a function of latency, as it depends not only on the model, but also on the hardware.
We note that we use a batch size of 1 in our setup (see \cref{app:base_and_quantised_llms}) and compute the corresponding token latency and throughput.}

{The throughput defines the number of output tokens that can be predicted per time unit. For sequential, non-parallel decoding, such as autoregressive LLM, latency and throughput are directly coupled. More specifically, for a batch size of 1, the throughput corresponds to $\frac{1}{\text{Latency}}$. Since decoding is inherently sequential and the setup is compute-bound, improving the latency directly improves throughput.}

%Therefore, across both stages, the overall latency depends primarily on the time required to load data from high-bandwidth memory into caches and registers, and the time required to perform the computations.On modern GPUs, loading a number from memory is often more than 100 times slower than performing an arithmetic operation on that number \citep{gao2019representation, dettmers2023case}. Consequently, memory access is the dominant cost. Therefore, reducing the time spent loading data from main memory can often accelerate the overall latency. Such reductions can be achieved mainly through caching and through the use of lower-precision numbers, which relate directly to the total model bits. 

{In \cref{tab:efficiency_metrics}, we compare the different efficiency metrics for models from the Qwen3 model family. Specifically, we select models with 4B, 8B, and 16B parameters. For the quantization, we use BitsandBytes \citep{dettmers_2022_bitsandbytes} and quantize the full-precision models to 4 and 8 bits. }

\newpage
\section{Reliability {Scaling Trends} Across Different Benchmarks}\label{app:additional_performance_scaling}

\begin{figure}[]
    \centering

    \begin{subfigure}[b]{0.48\textwidth}
        \centering
        \includegraphics[width=\textwidth]{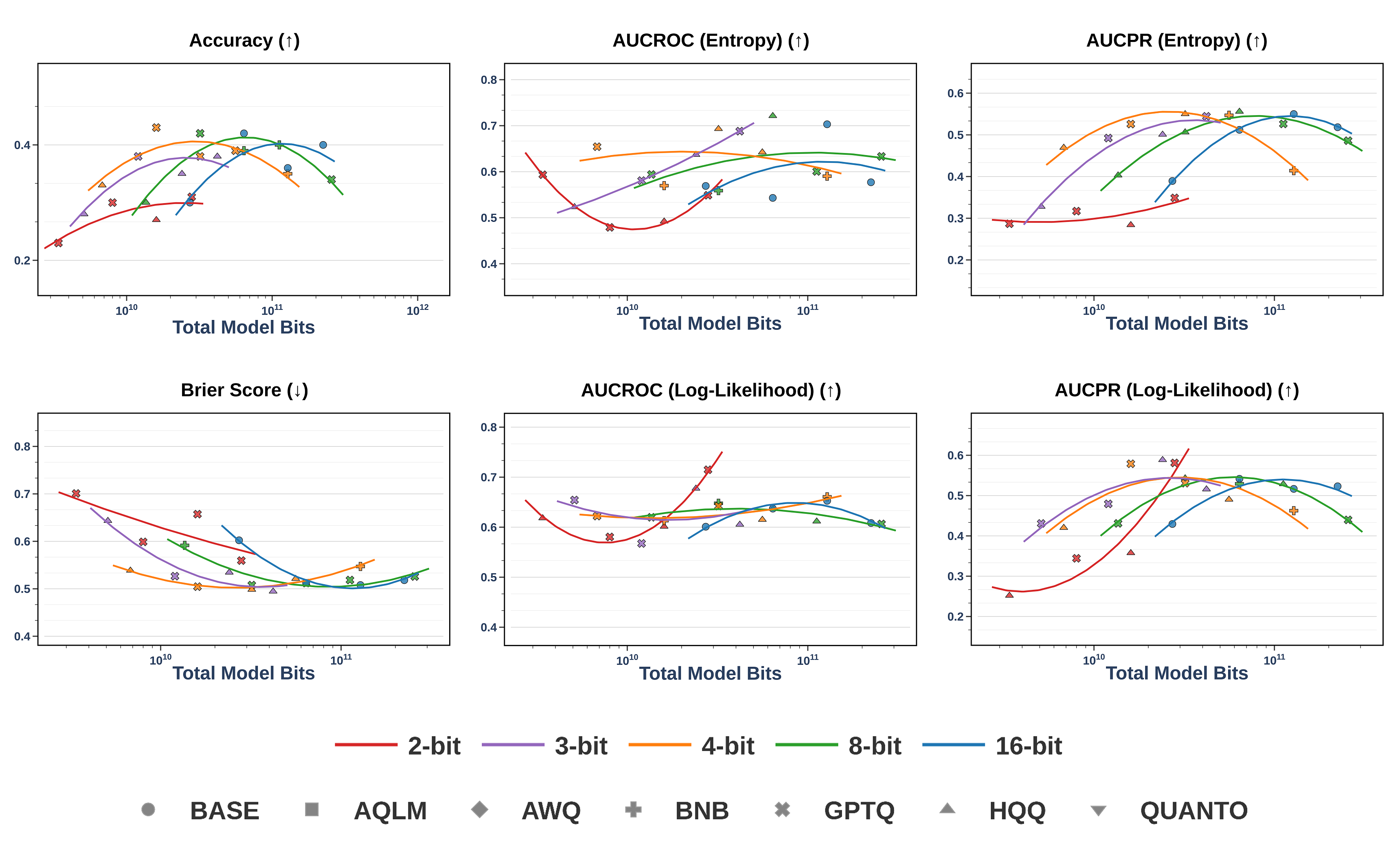}
        \caption{RACE \citep{lai2017race}}
        \label{fig:sub_race_qwen}
    \end{subfigure}
    \hfill
    \begin{subfigure}[b]{0.48\textwidth}
        \centering
        \includegraphics[width=\textwidth]{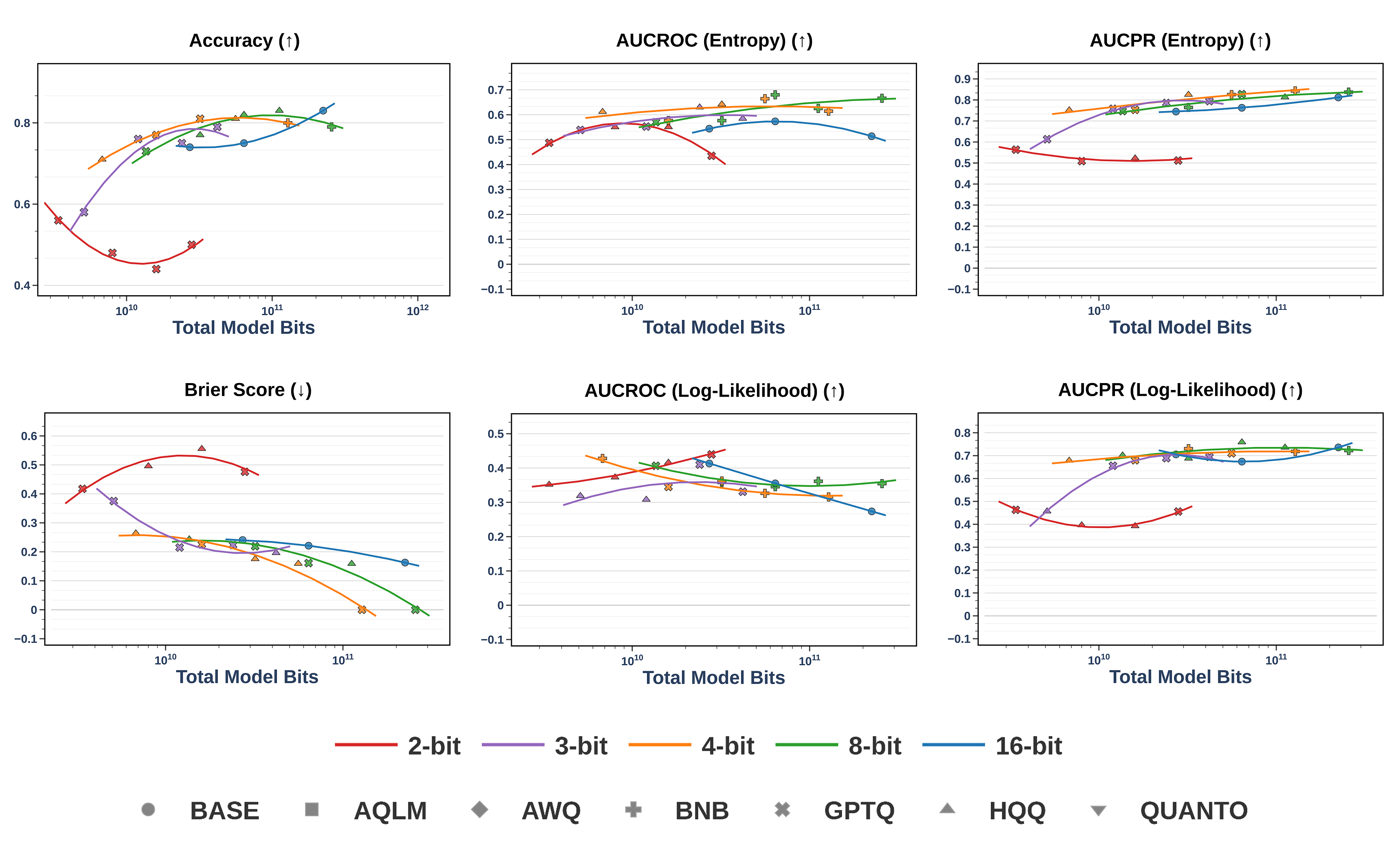}
        \caption{PiQA \citep{bisk2020piqa}}
        \label{fig:sub_piqa_qwen}
    \end{subfigure}

    \vskip 0.3cm

    \begin{subfigure}[b]{0.48\textwidth}
        \centering
        \includegraphics[width=\textwidth]{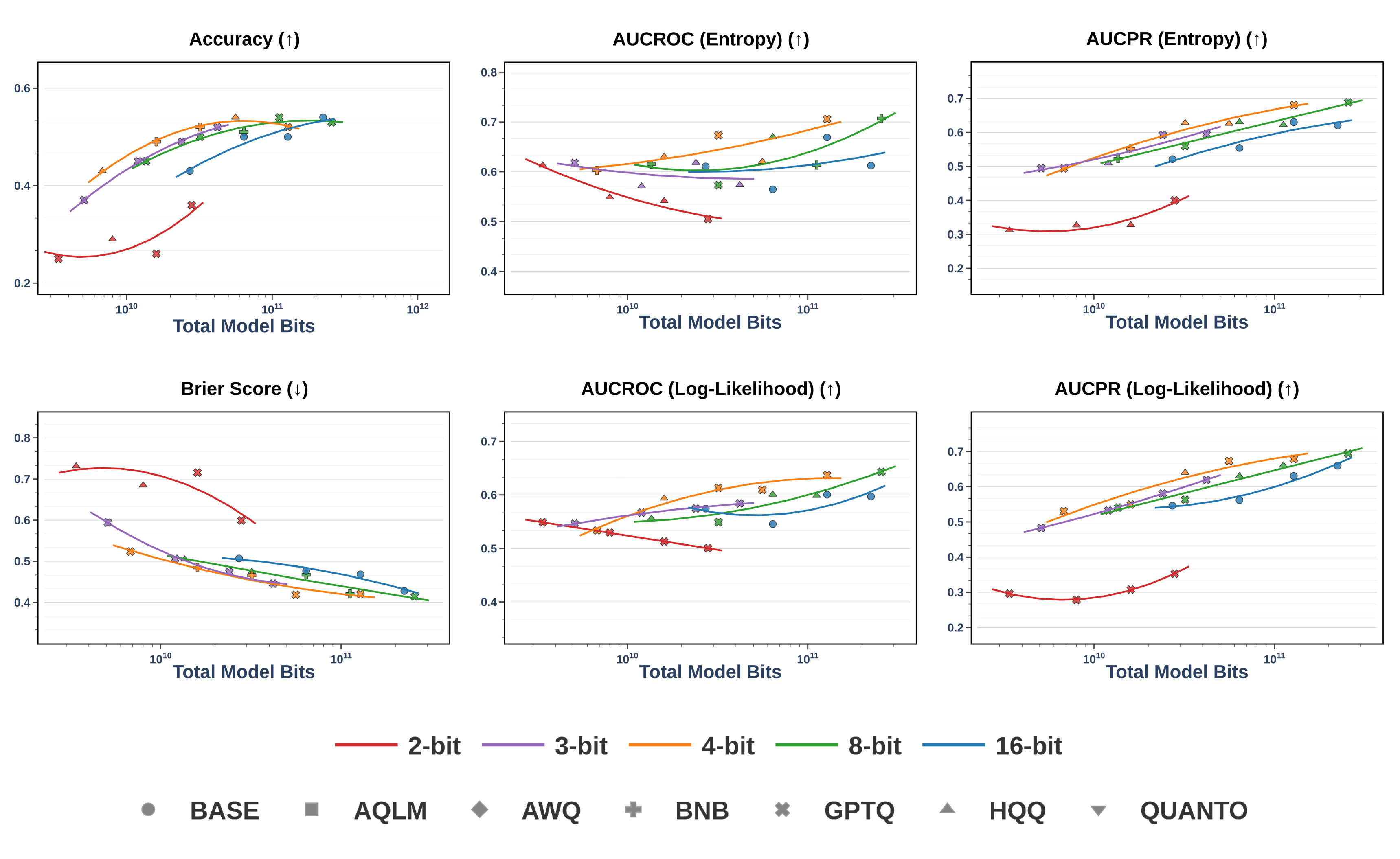}
        \caption{HellaSwag \citep{zellers2019hellaswag}}
        \label{fig:sub_hellaswag_qwen}
    \end{subfigure}
    \hfill
    \begin{subfigure}[b]{0.48\textwidth}
        \centering
        \includegraphics[width=\textwidth]{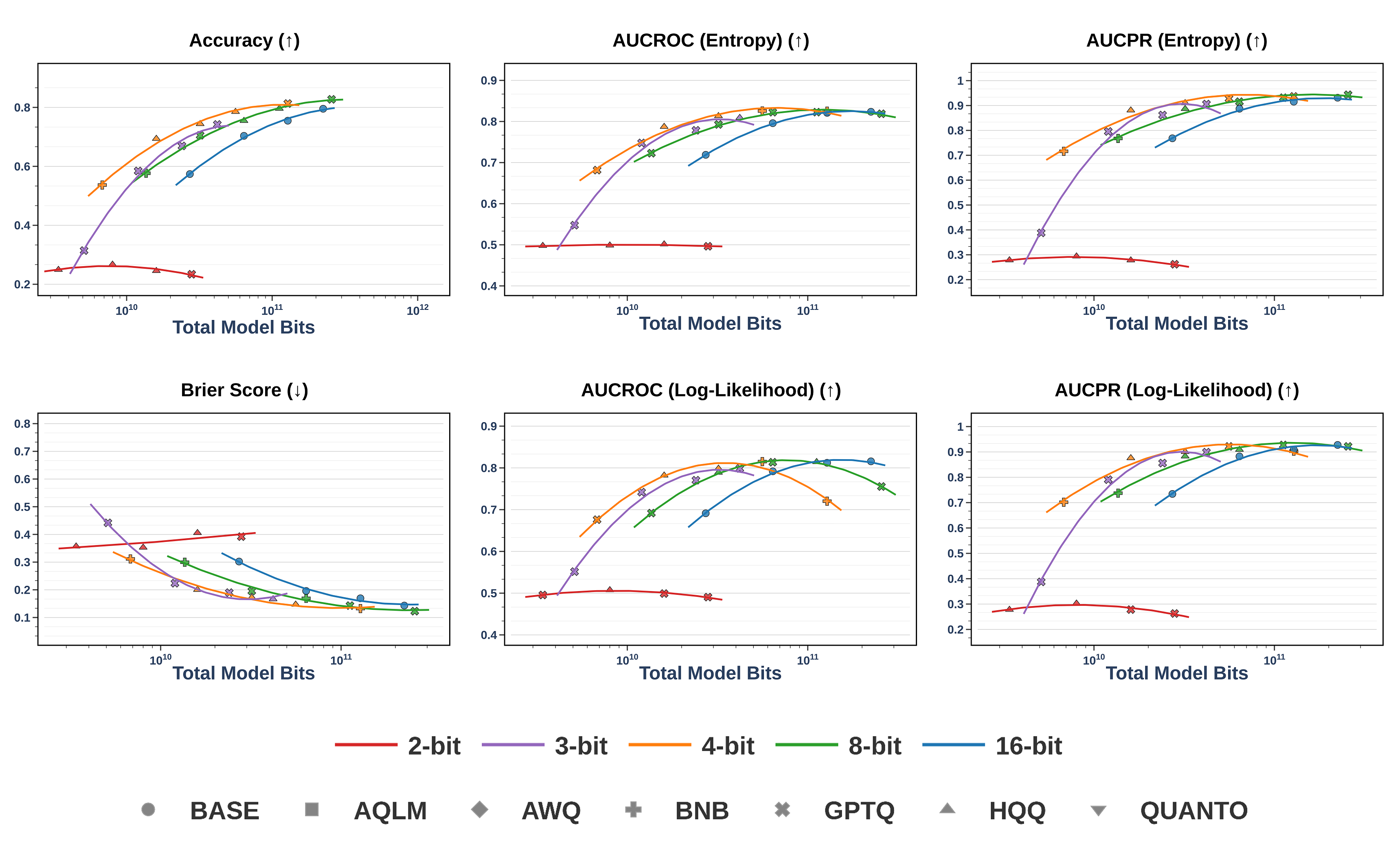}
        \caption{MMLU \citep{hendrycks2020measuring}}
        \label{fig:sub_mmlu_qwen}
    \end{subfigure}
    \vspace{1em}
    \caption{Bit-level scaling trends of Qwen3 models across four benchmarks.}
    \label{fig:qwen_bit_scalings_all}
\end{figure}

\begin{figure}[]
    \centering

    \begin{subfigure}[b]{0.48\textwidth}
        \centering
        \includegraphics[width=\textwidth]{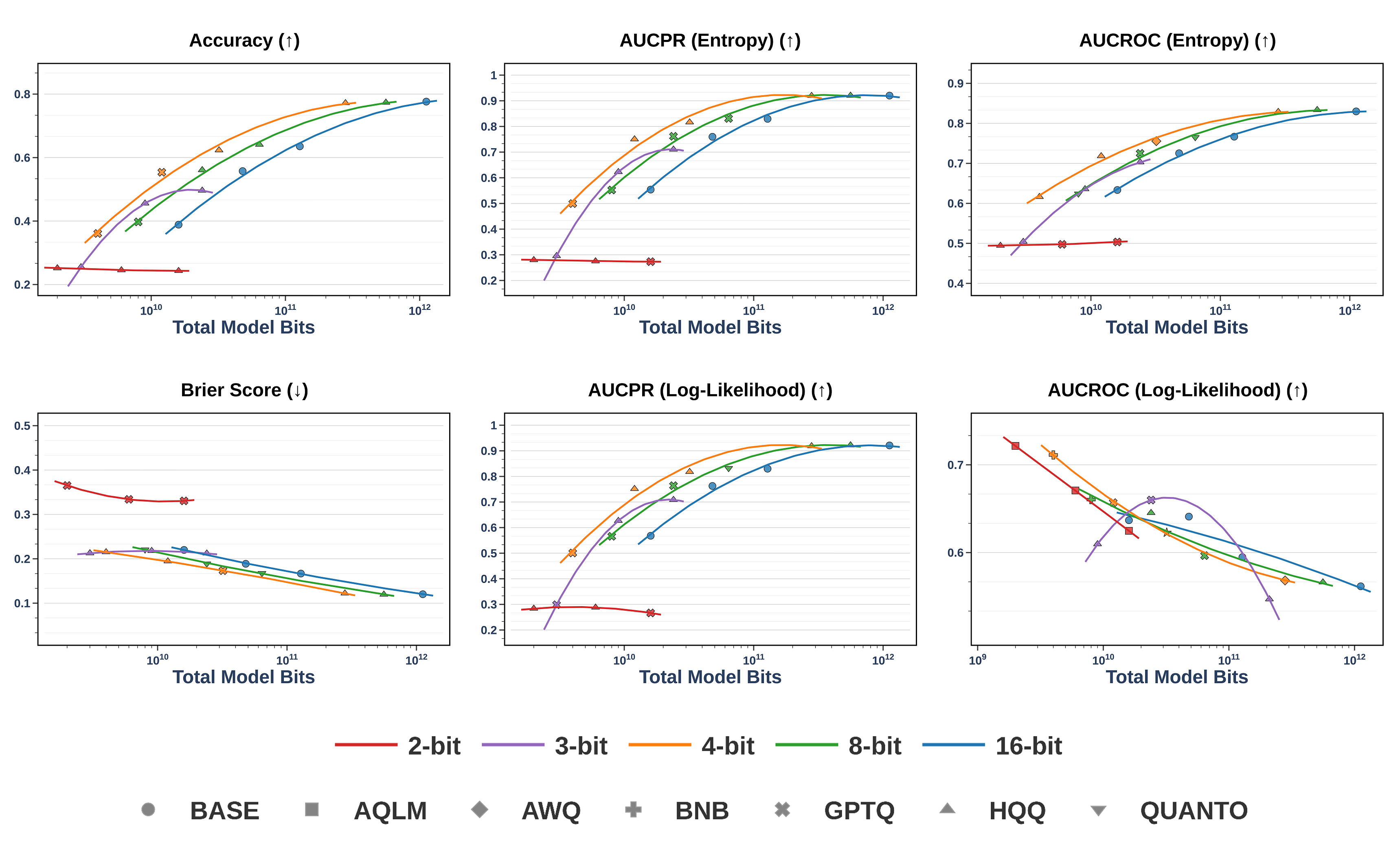}
        \caption{MMLU \citep{hendrycks2020measuring}}
        \label{fig:sub_mmlu}
    \end{subfigure}
    \hfill
    \begin{subfigure}[b]{0.48\textwidth}
        \centering
        \includegraphics[width=\textwidth]{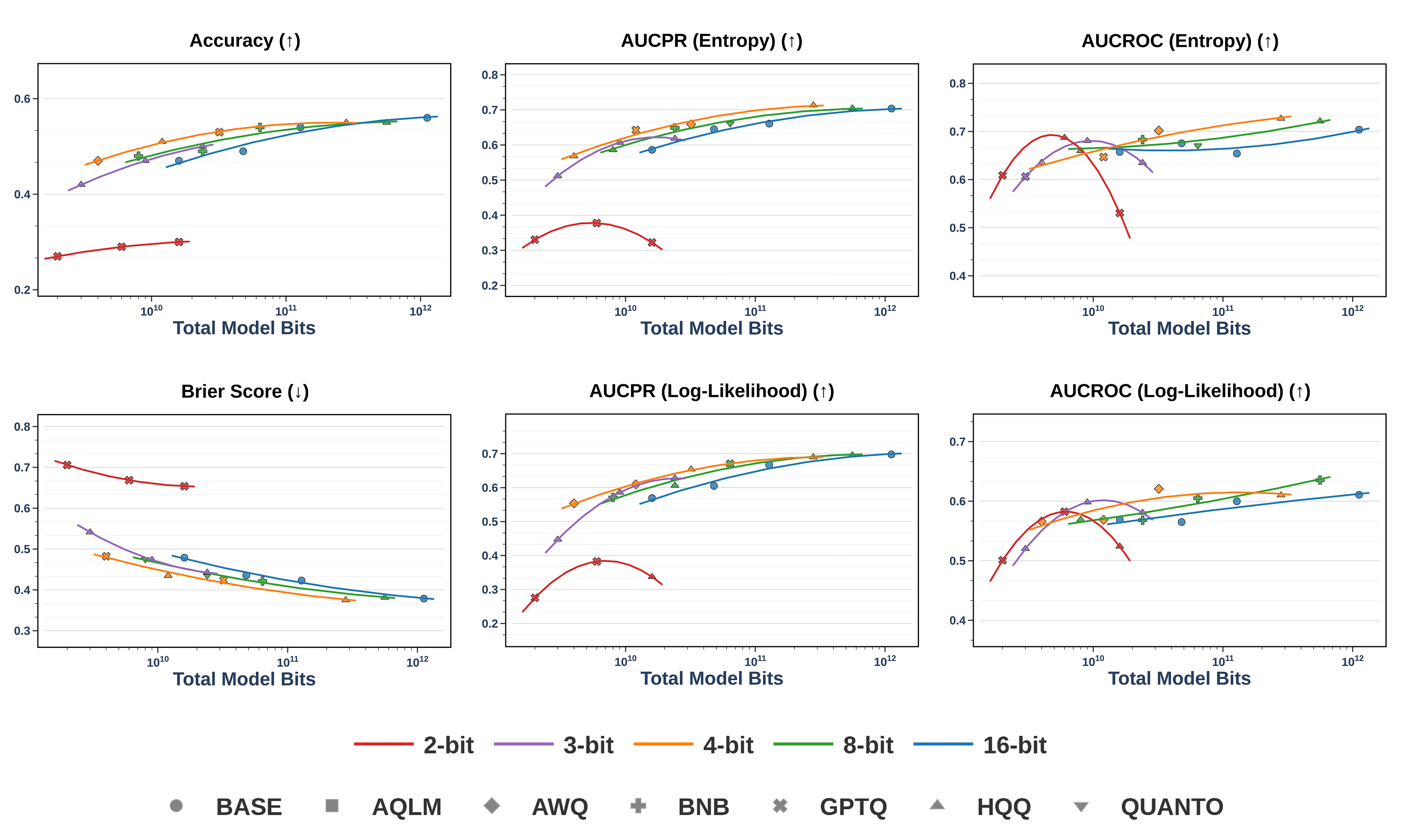}
        \caption{HellaSwag \citep{zellers2019hellaswag}}
        \label{fig:sub_hellaswag}
    \end{subfigure}

    \vskip 0.3cm

    \begin{subfigure}[b]{0.48\textwidth}
        \centering
        \includegraphics[width=\textwidth]{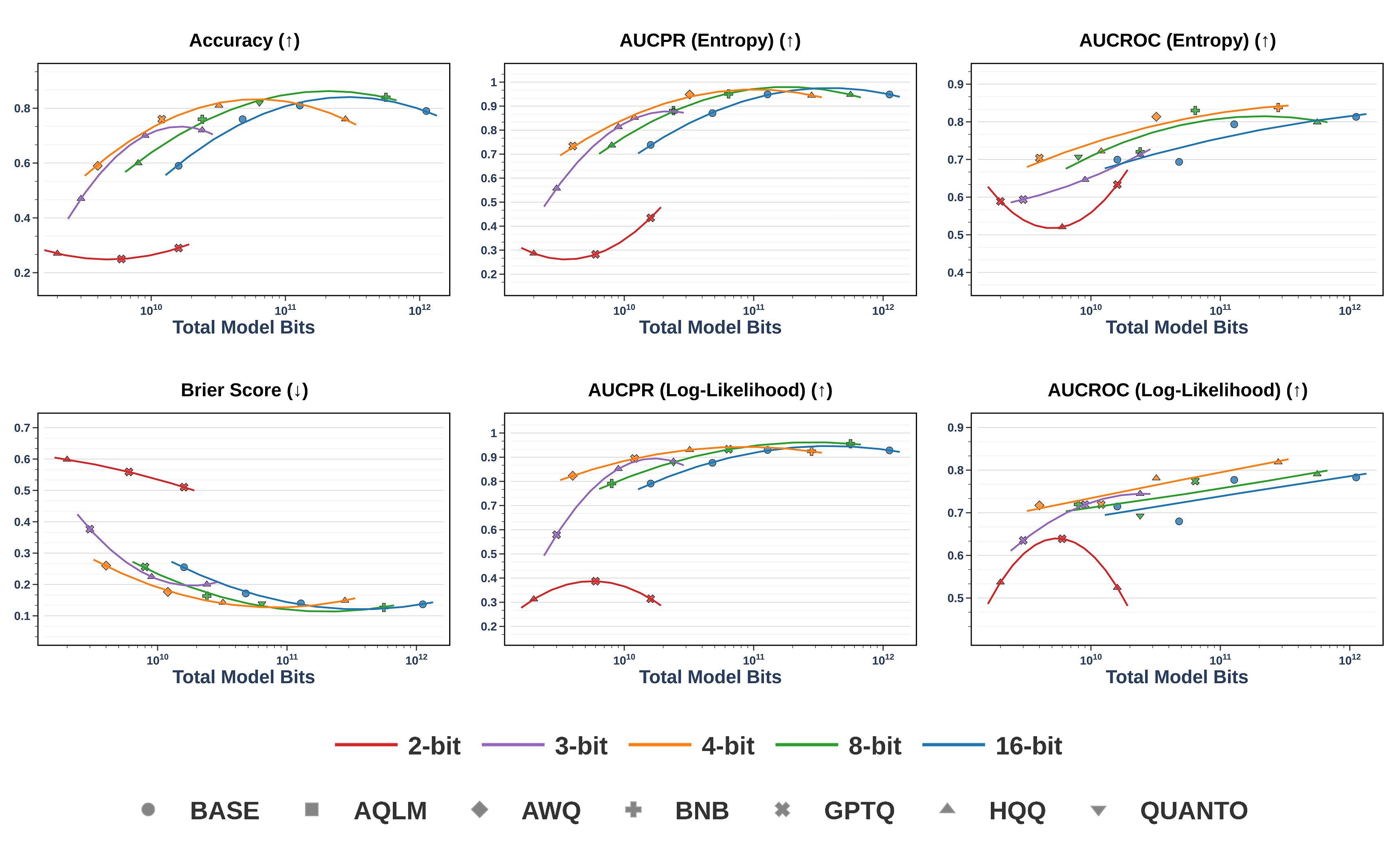}
        \caption{Arc Easy \citep{clark2018think}}
        \label{fig:sub_arc_easy}
    \end{subfigure}
    \hfill
    \begin{subfigure}[b]{0.48\textwidth}
        \centering
        \includegraphics[width=\textwidth]{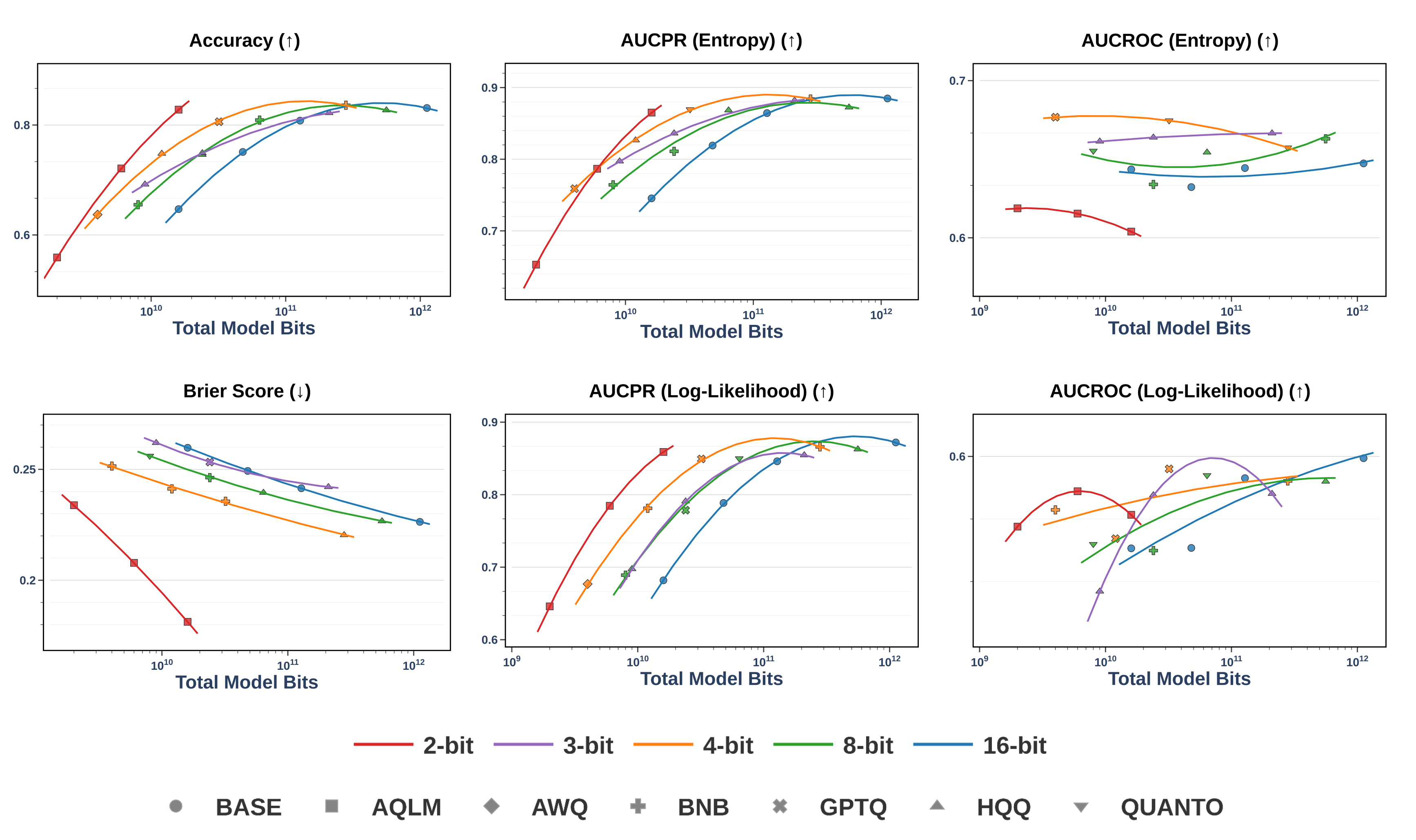}
        \caption{CoQA \citep{reddy2019coqa}}
        \label{fig:sub_coqa}
    \end{subfigure}
    \vspace{1em}
    \caption{Bit-level scaling trends of LLaMA models across four benchmarks.}
    \label{fig:llama_bit_scalings_all}
\end{figure}

\paragraph{{Reliability scaling trends of Qwen3 models}}
{We evaluate four base models from the Qwen3 model family, with sizes of 4B, 8B, 14B, and 32B, and their corresponding quantized models using 6 different PTQ methods. We provide results in \cref{fig:qwen_bit_scalings_all}}  
{We evaluate the reasoning abilities on the PiQA benchmark in \cref{fig:sub_piqa_qwen}, and the understanding abilities on RACE in \cref{fig:sub_race_qwen}. For instruction-following assessment, we use the HellaSwag benchmark and provide results in \cref{fig:sub_hellaswag}. For in-context learning, we evaluate on the widely-used MMLU dataset in \cref{fig:sub_mmlu_qwen}. Across different datasets, model sizes, and bitwidths, we observe a sweet spot at 4-bit quantization. Both GPTQ and HQQ provide the best performance for 4-bit quantization. We further note that for Qwen3 models, 3-bit quantization can provide favorable performance in terms of zero-shot performance and reliability. }

\begin{figure}[]
 %\vspace{-1em}
    \centering    \includegraphics[width=0.8\textwidth]{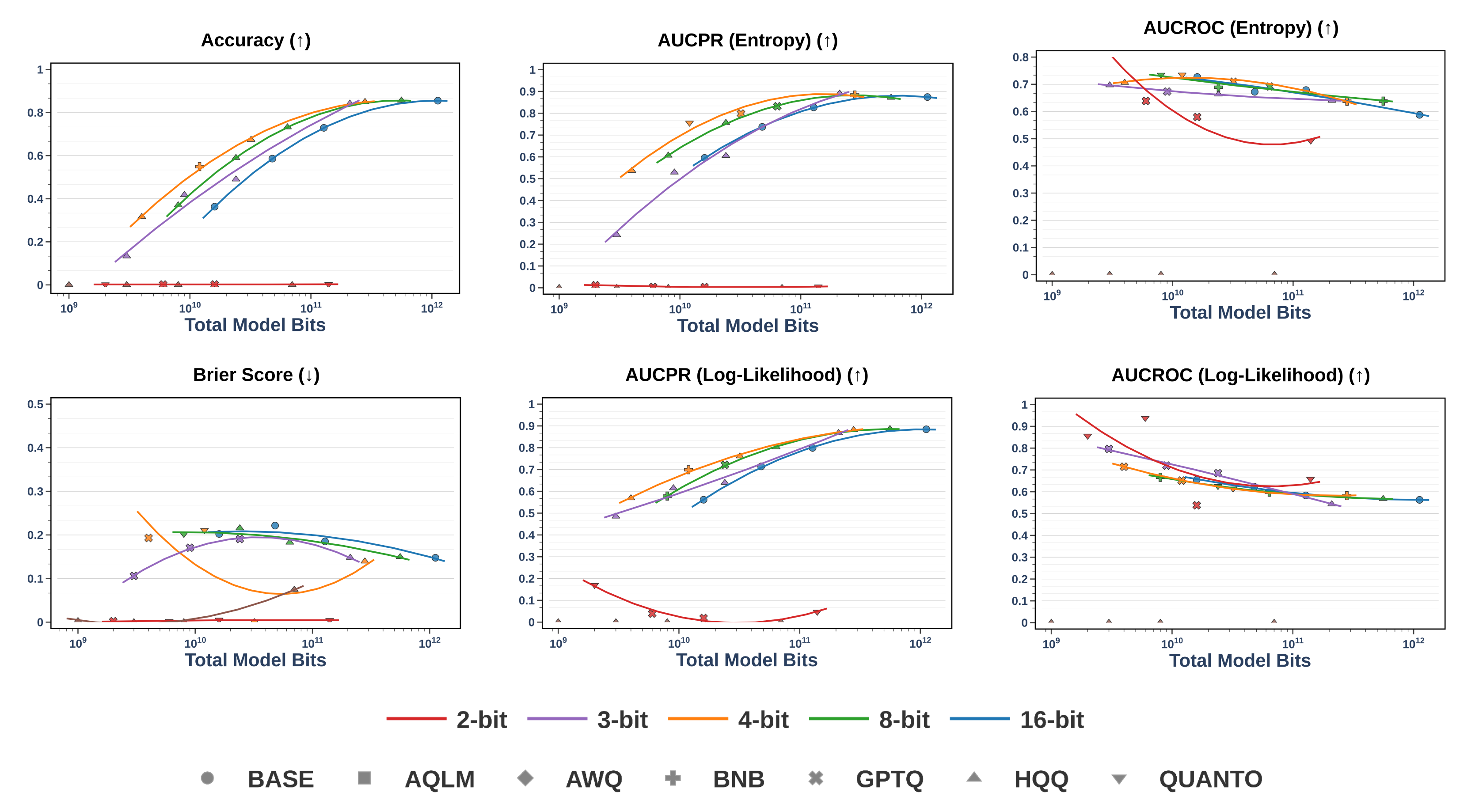}
    \caption{Scaling behavior of the reliability metrics of the \textbf{instruction-tuned} LLaMA models on \textbf{TriviaQA}.} 
    \label{fig:app_instruct_llama_triviaqa}
    %\vspace{-1em}
\end{figure}

\paragraph{{Reliability scaling trends of LLaMA-3 models}}
{
In addition to the performance and reliability bit-level inference scalings provided in \cref{sec:results}, we extend the evaluation to in-context learning and instruction-following tasks to assess the emergent abilities of quantized LLMs from the LLaMA-3 model family.
For in-context learning, we provide the bit-level inference scaling plots under \cref{fig:sub_mmlu} for the MMLU task. For instruction-following, we evaluate on HellaSwag in \cref{fig:sub_hellaswag} and ARC in \cref{fig:sub_arc_easy}. We further assess the open-ended generation on CoQA in \cref{fig:sub_coqa}, which is a conversational question-answering task that tests the dialog understanding capabilities of models. Additional bit-level scalings on the instruction-tuned LLaMA models are presented in \cref{fig:app_instruct_llama_triviaqa}.} For the 2-bit precision, AQLM-PV provides the best zero-shot accuracy. For 3-bit models, HQQ outperforms GPTQ across different metrics. For 4-bit models, both AWQ \citep{lin2024awq} and GPTQ \citep{frantar2022gptq} generally provide the best downstream task performance and reliability. 

%\begin{figure}[]
 %\vspace{-1em}
%    \centering    \includegraphics[width=0.8\textwidth]{figures/appendix_reliability_log_quadr_MMLU.png}
%    \caption{Bit-level scaling trends of LLaMA models on the \textbf{MMLU} \citep{hendrycks2020measuring} benchmark.} 
%    \label{fig:app_mmlu_scalings}
    %\vspace{-1em}
%\end{figure}

\newpage
\section{{Reliability} Scaling Trends on perturbed {Benchmarks}}\label{sec:additional_perturbation_scalings}
In this section, we provide additional results to show the scaling behavior of the accuracy, the quality of the uncertainty estimates, the log-likelihood, and the calibration using the perturbed prompts. For all experiments, we fit a log-quadratic function per bit width, and only show the best-performing model for every bit level. We further report the mean and standard deviation of the accuracy and the reliability metrics of the four base LLaMA-3 models, under unperturbed and perturbed input prompts, in \cref{tab:best_methods_per_bit_width}. Additionally, in \cref{tab:best_methods_per_bit_width_intensity_4}, we provide a list of the best quantization methods per base model and precision for two different perturbation intensities. For the 2-bit precision, AQLM-PV provides the best performance on the perturbed datasets. For 3-bit models, HQQ outperforms GPTQ across different metrics. For 4-bit models, HQQ \citep{badri_half-quadratic_nodate} and GPTQ \citep{frantar2022gptq} generally provide the best downstream performance and reliability under perturbation, followed by AWQ \citep{lin2024awq}.
\begin{table}[]
\centering
\caption{Performance across different perturbation intensities on the TriviaQA dataset}
\label{tab:best_methods_per_bit_width}
\begin{small}
\begin{tabular}{llccc}
\toprule
\multirow{2}{*}{{Model}} & \multirow{2}{*}{{Intensity}} & \multicolumn{3}{c}{{Metrics}} \\
\cmidrule(l){3-5}
 & & {Accuracy} & {AUCROC (Entropy)} & {AUCPR (Entropy)} 
 %& {Brier Score} 
 \\
\midrule
\multirow{3}{*}{{LLaMA-3.2-1B}} & 0  & $0.338 \pm 0.000$ & $0.683 \pm 0.000$ & $0.518 \pm 0.000$ 
%& $0.201 \pm 0.000$ 
\\
  & 4  & $0.192 \pm 0.050$ & $0.636 \pm 0.029$ & $0.264 \pm 0.084$ 
  %& $0.172 \pm 0.015$ 
  \\
  & 16  & $0.078 \pm 0.077$ & $0.578 \pm 0.069$ & $0.098 \pm 0.118$ 
  %& $0.156 \pm 0.046$ 
  \\
\midrule
\multirow{3}{*}{{LLaMA-3.2-3B}} & 0  & $0.549 \pm 0.000$ & $0.722 \pm 0.000$ & $0.746 \pm 0.000$ 
%& $0.222 \pm 0.000$ 
\\
  & 4  & $0.384 \pm 0.064$ & $0.689 \pm 0.020$ & $0.534 \pm 0.083$ 
  %& $0.213 \pm 0.012$ 
  \\
  & 16  & $0.189 \pm 0.126$ & $0.608 \pm 0.076$ & $0.234 \pm 0.170$ 
  %& $0.199 \pm 0.041$ 
  \\
\midrule
\multirow{3}{*}{{LLaMA-3-8B}} & 0  & $0.691 \pm 0.000$ & $0.715 \pm 0.000$ & $0.847 \pm 0.000$ 
%& $0.222 \pm 0.000$
\\
  & 4  & $0.519 \pm 0.069$ & $0.691 \pm 0.022$ & $0.687 \pm 0.065$ 
  %& $0.230 \pm 0.012$ 
  \\
  & 16  & $0.283 \pm 0.156$ & $0.640 \pm 0.051$ & $0.357 \pm 0.189$ 
  %& $0.202 \pm 0.034$ 
  \\
\midrule
\multirow{3}{*}{{LLaMA-3-70B}} & 0  & $0.728 \pm 0.000$ & $0.503 \pm 0.000$ & $0.729 \pm 0.000$ 
%& $0.277 \pm 0.000$ 
\\
  & 4  & $0.706 \pm 0.037$ & $0.557 \pm 0.036$ & $0.736 \pm 0.036$ 
  %& $0.301 \pm 0.014$ 
  \\
  & 16  & $0.503 \pm 0.145$ & $0.595 \pm 0.063$ & $0.544 \pm 0.136$ 
  %& $0.272 \pm 0.037$ 
  \\
\bottomrule
\end{tabular}
\end{small}
\end{table}
\begin{table}[]
\centering
\caption{Recommendation list of the best quantization methods per model size and bit width across different evaluation metrics on the unperturbed TriviaQA dataset.}
\begin{small}
\begin{tabular}{cc|cccc}
\toprule
% &  & \multicolumn{4}{c}{{Best Method}} \\
%\cmidrule(l){3-6}
\multirow{2}{*}{\shortstack{Base \\ Model}} & \multirow{2}{*}{\shortstack{Bit \\ Width}}
&  Accuracy& {AUCROC} & {AUCPR} & {AUCPR} \\
& & & (Entropy) & (Entropy) & {(Log-Lik.)} \\
\toprule
1B & 2 & AQLM & AQLM & AQLM & AQLM \\
1B & 4 & AWQ & GPTQ & AWQ & GPTQ \\
1B & 8 & GPTQ & Quanto & GPTQ & BNB \\
%1B & 16 & Base & Base & Base & Base \\
\midrule
3B & 2 & AQLM & AQLM & AQLM & AQLM \\
3B & 3 & HQQ & HQQ & HQQ & HQQ \\
3B & 4 & HQQ & GPTQ & HQQ & GPTQ \\
3B & 8 & Quanto & BNB & Quanto & Quanto \\
%3B & 16 & Base & Base & Base & Base \\
\midrule
8B & 2 & AQLM & AQLM & AQLM & AQLM \\
8B & 3 & HQQ & HQQ & HQQ & HQQ \\
8B & 4 & HQQ & GPTQ & AWQ & HQQ \\
8B & 8 & HQQ & Quanto & HQQ & GPTQ \\
%8B & 16 & Base & Base & Base & Base \\
\midrule
70B & 3 & HQQ & HQQ & HQQ & HQQ \\
70B & 4 & AWQ & AWQ & AWQ & AWQ \\
70B & 8 & HQQ & BNB & Quanto & HQQ \\
%70B & 16 & Base & Base & Base & Base \\
\bottomrule
\end{tabular}
\end{small}
\label{tab:best_methods_intensity_0_table}
\end{table}
\begin{table}[]
\centering
\caption{Recommendation list of the best quantization methods per model size and bit width across different evaluation metrics on the \textbf{perturbed} TriviaQA dataset. We average the performance over all 15 perturbations with an intensity equal to \textbf{4}.}
\begin{small}
\begin{tabular}{cc|cccc}
\toprule
% &  & \multicolumn{4}{c}{{Best Method}} \\
%\cmidrule(l){3-6}
\multirow{2}{*}{\shortstack{Base \\ Model}} & \multirow{2}{*}{\shortstack{Bit \\ Width}}
&  Accuracy& {AUCROC} & {AUCPR} & {AUCPR} \\
& & & (Entropy) & (Entropy) & {(Log-Lik.)} \\
\toprule
1B & 2 & AQLM & AQLM & AQLM & - \\
1B & 4 & AWQ & GPTQ & AWQ & HQQ \\
1B & 8 & HQQ & GPTQ & HQQ & BNB \\
%1B & 16 & Base & Base & Base & Base \\
\midrule
3B & 2 & AQLM & AQLM & AQLM & - \\
3B & 3 & HQQ & HQQ & HQQ & HQQ \\
3B & 4 & HQQ & GPTQ & HQQ & Quanto \\
3B & 8 & Quanto & BNB & Quanto & HQQ \\
%3B & 16 & Base & Base & Base & Base \\
\midrule
8B & 2 & AQLM & AQLM & AQLM & - \\
8B & 3 & HQQ & HQQ & HQQ & HQQ \\
8B & 4 & AWQ & GPTQ & AWQ & HQQ \\
8B & 8 & HQQ & Quanto & GPTQ & Quanto \\
%8B & 16 & Base & Base & Base & Base \\
\midrule
70B & 3 & HQQ & HQQ & HQQ & HQQ \\
70B & 4 & AWQ & AWQ & AWQ & HQQ \\
70B & 8 & HQQ & BNB & HQQ & HQQ \\
%70B & 16 & Base & Base & Base & Base \\
\bottomrule
\end{tabular}
\end{small}
\label{tab:best_methods_per_bit_width_intensity_4}
\end{table}

\newpage
\section{{Additional Ablations}}
\begin{figure}[]
 \vspace{-0.5em}
    \centering
    \includegraphics[width=0.5\textwidth]{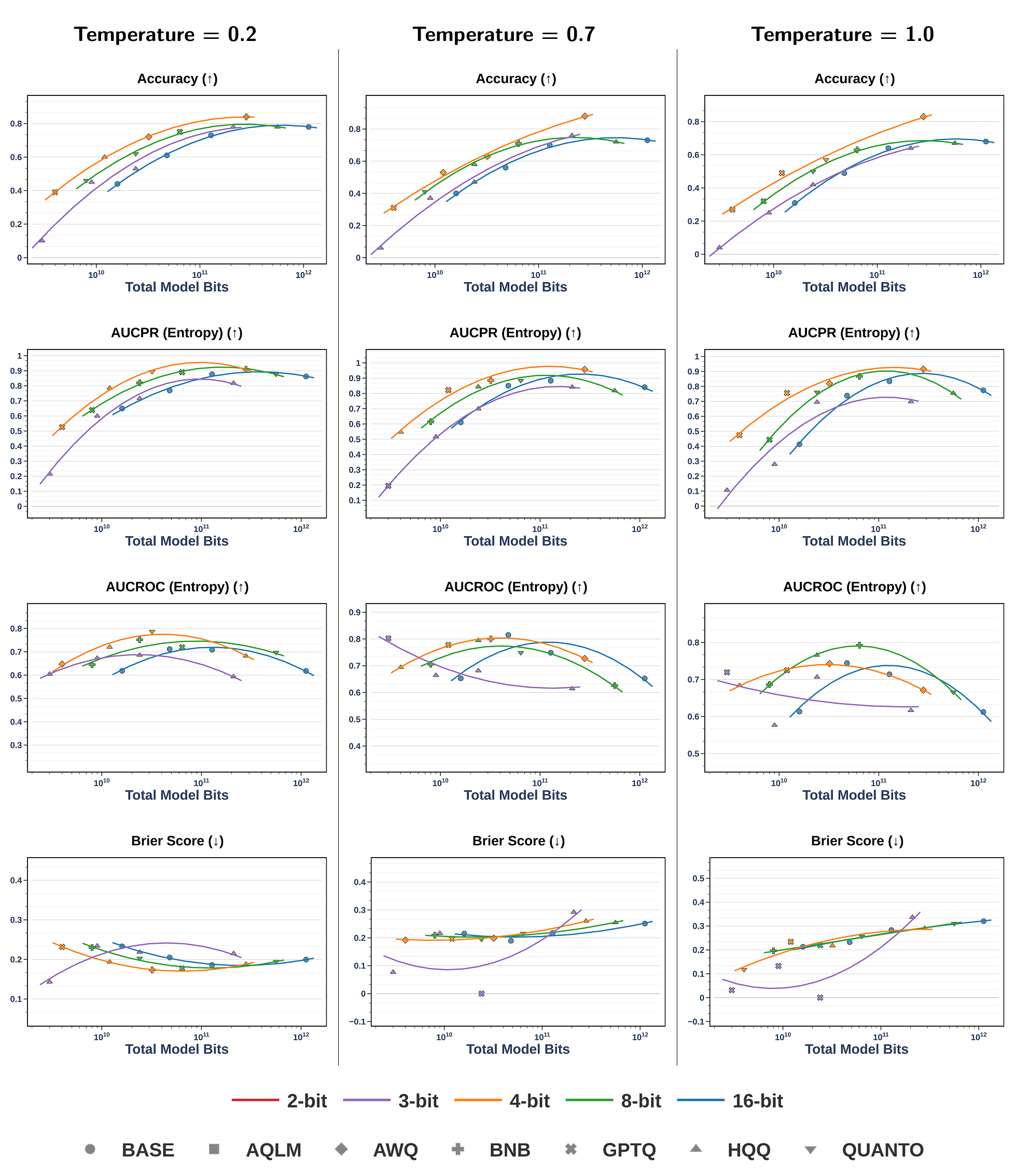}
    \caption{{Bit-level inference scaling trends on TriviaQA for various temperature values for sampling.}} 
    \label{fig:temperature_ablation}
    \vspace{-0.5em}
\end{figure}

\begin{figure}[]
 \vspace{-0.5em}
    \centering
    \includegraphics[width=0.5\textwidth]{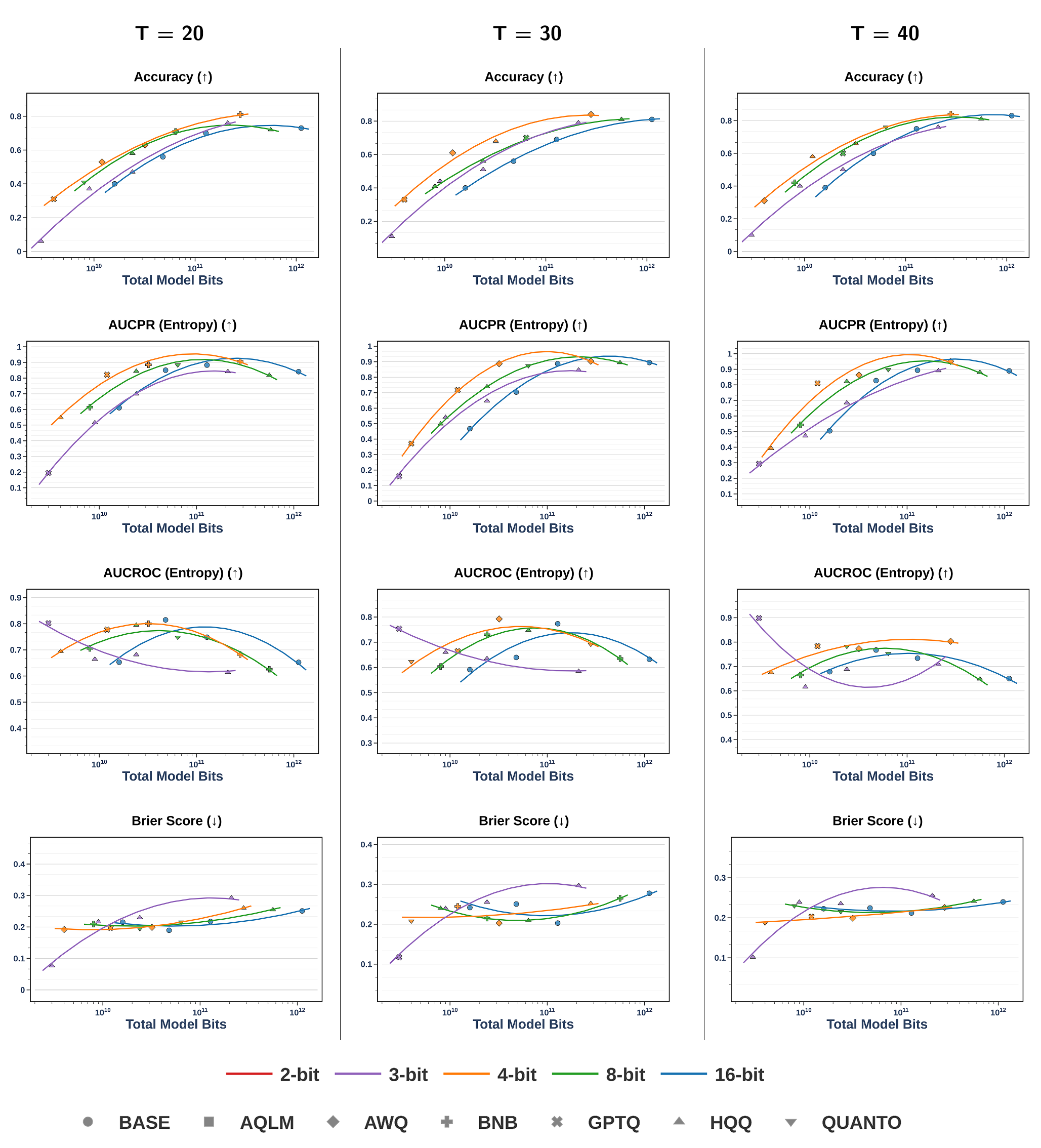}
    \caption{{Bit-level inference scaling trends on TriviaQA for different numbers of output tokens.}} 
    \label{fig:output_token_ablations}
    \vspace{-0.5em}
\end{figure}

{In this section, we provide additional ablations across temperature used for sampling and the length of the generated sequences. We note that we limit the evaluation to 100 samples from the TriviaQA dataset.}
{
We first examine the bit-level inference scalings under different decoding strategies. Specifically, evaluate the open-ended generation of base and quantized LLMs on TriviaQA, where we sample with various temperature values in {0.2, 0.7, 1.0}. We present the qualitative results in \cref{fig:temperature_ablation}.
Overall, the performance and reliability metrics exhibit consistent trends across temperature settings.
In particular, accuracy increases with the total number of model bits, with 4-bit models achieving the strongest performance.
For reliability, we observe a pronounced peak for 4-bit quantized models. However, as the temperature increases to 0.7 and 1.0, 3-bit quantized models yield the best Brier scores, given a fixed total model bits.} {In \cref{fig:radar_triviaqa}, we truncate the generation of models to 20 tokens. We explore longer generations using an increased number of tokens. We present results in \cref{fig:output_token_ablations}.}

\section{{How does Quantization-Aware Training affect the scaling trends?}}\label{sec:QAT}
{While our study primarily focuses on state-of-the-art post-training quantization techniques for LLMs, we investigate in the following how quantization-aware training (QAT) influences the performance and reliability scaling trends. QAT is especially compelling for extreme quantization regimes, as post-training quantizations often result in substantial performance degradation.}

\begin{figure}[ht]
 \vspace{-0.5em}
    \centering
    \includegraphics[width=0.5\textwidth]{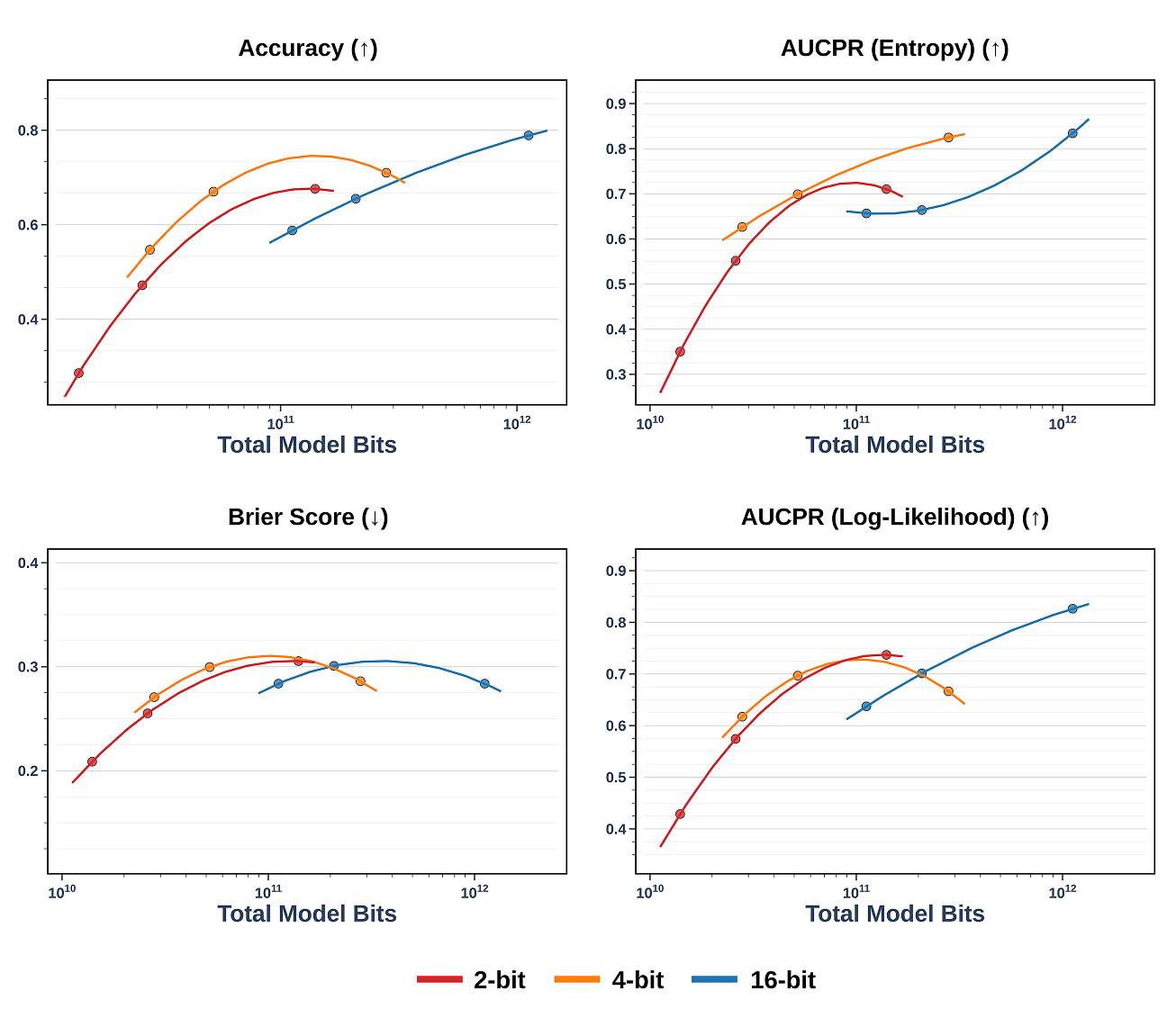}
    \caption{{Bit-level scaling trends of quantized LLaMA models using EfficientQAT \citep{chen2025efficientqat}.}} 
    \label{fig:efficientqat}
    \vspace{-1em}
\end{figure}

{
In the following, we examine the recent EfficientQAT \citep{chen2025efficientqat} method, which consists of two stages: block-wise training of all model parameters followed by end-to-end training of the quantization parameters. We apply QAT to three LLaMA-3 models with 7B, 13B, and 70B parameters and evaluate 4-bit and 2-bit quantized models as well as their full-precision counterparts.} 
{We present qualitative results of the bit-level inference scalings in \cref{fig:efficientqat}. We find that 4-bit quantized models generally offer favorable zero-shot performance and reliability for a fixed total bit budget. 
EfficientQAT \citep{chen2025efficientqat} achieves impressive results in 2-bit scenarios, improving the zero-shot performance and reliability scaling trends compared to PTQ approaches. However, a performance gap remains compared to higher bitwidths.
}

{
We limit the evaluation to the EfficientQAT approach due to time and resource constraints. Extending the bit-level reliability scaling study to different QAT approaches \citep{liu2024llm, ma2024era,xu2024onebit} and performing a more thorough comparison across different model backbones, tasks, and bit precisions could yield novel insights, which we leave for future work.
While QAT techniques are promising for extremely low bit-widths, there remains a need for more efficient and practical approaches. For example, while BitNet b1.58 \citep{ma2024era} achieves nearly lossless ternary quantization, it requires retraining LLMs from scratch on the entire pre-trained dataset. This makes it impractical for huge models and restricts its validation to 3B models trained on 100B tokens, limiting its applicability for comprehensive scaling law studies across varying model sizes.
}

\newpage
\section{{Are the bit-level inference scalings consistent across different quantization methods?}}

{To assess whether the observed bit-level inference scaling is specific to a single quantization method, we compare several 4-bit PTQ techniques across different model families. In particular, we evaluate HQQ, GPTQ, BNB, AWQ, and Quanto, and compare them against the corresponding 16-bit full-precision baselines. The results show that the overall trends are largely consistent across quantization methods: performance improves with the total number of model bits, while reliability-related metrics exhibit non-linear trends. Importantly, the 4-bit models often match or outperform the full-precision baselines in calibration and uncertainty-based reliability metrics, suggesting that the reliability peak is not tied to a specific quantization method.}

\begin{figure}[ht]
 \vspace{-0.5em}
    \centering
    \includegraphics[width=0.8\textwidth]{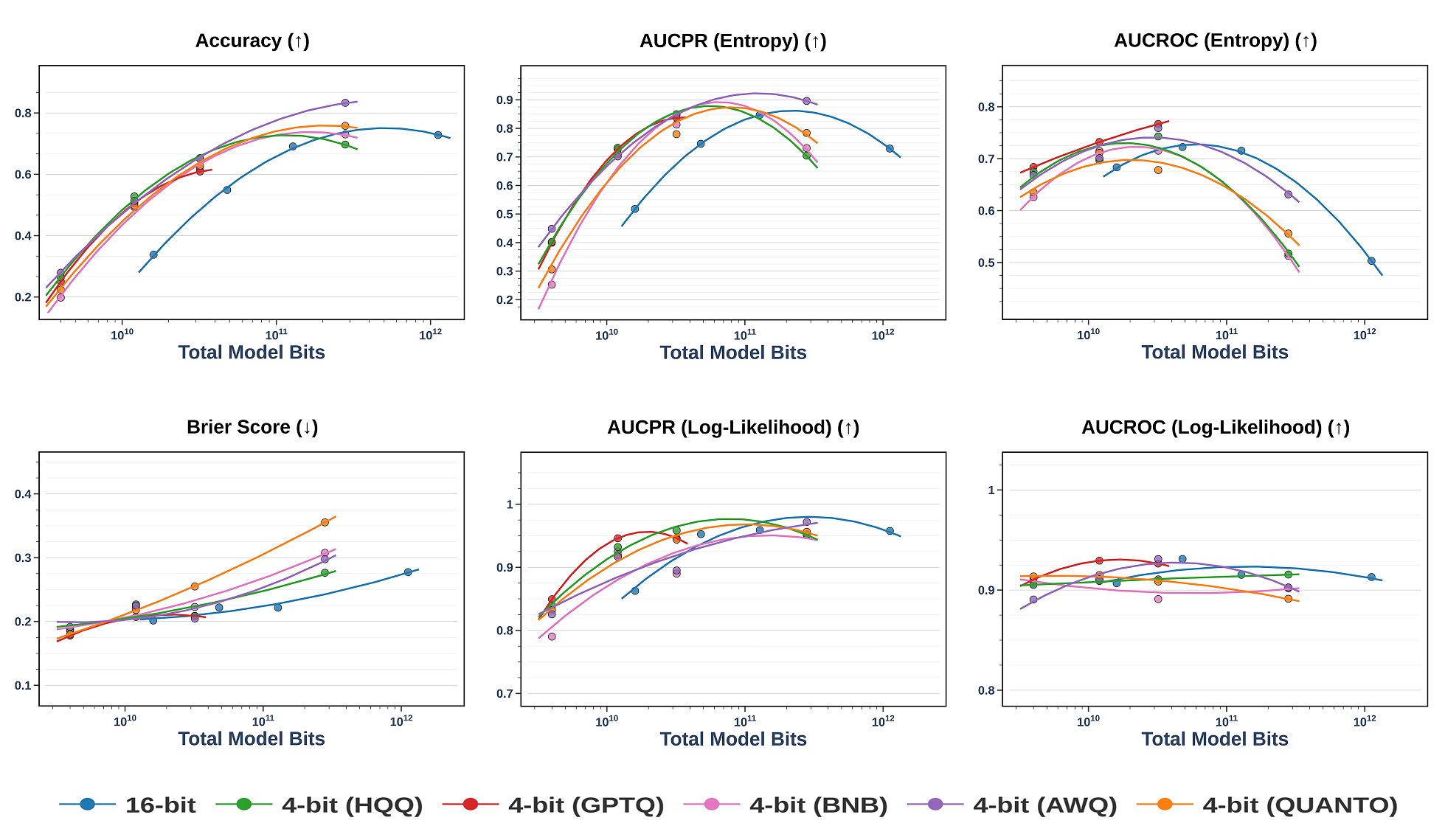}
    \caption{{Bit-level inference scaling trends of LLaMA models under different 4-bit quantization methods.}} 
    \label{fig:llama_4bit}
    \vspace{-1em}
\end{figure}
\begin{figure}[ht]
 \vspace{-0.5em}
    \centering
    \includegraphics[width=0.8\textwidth]{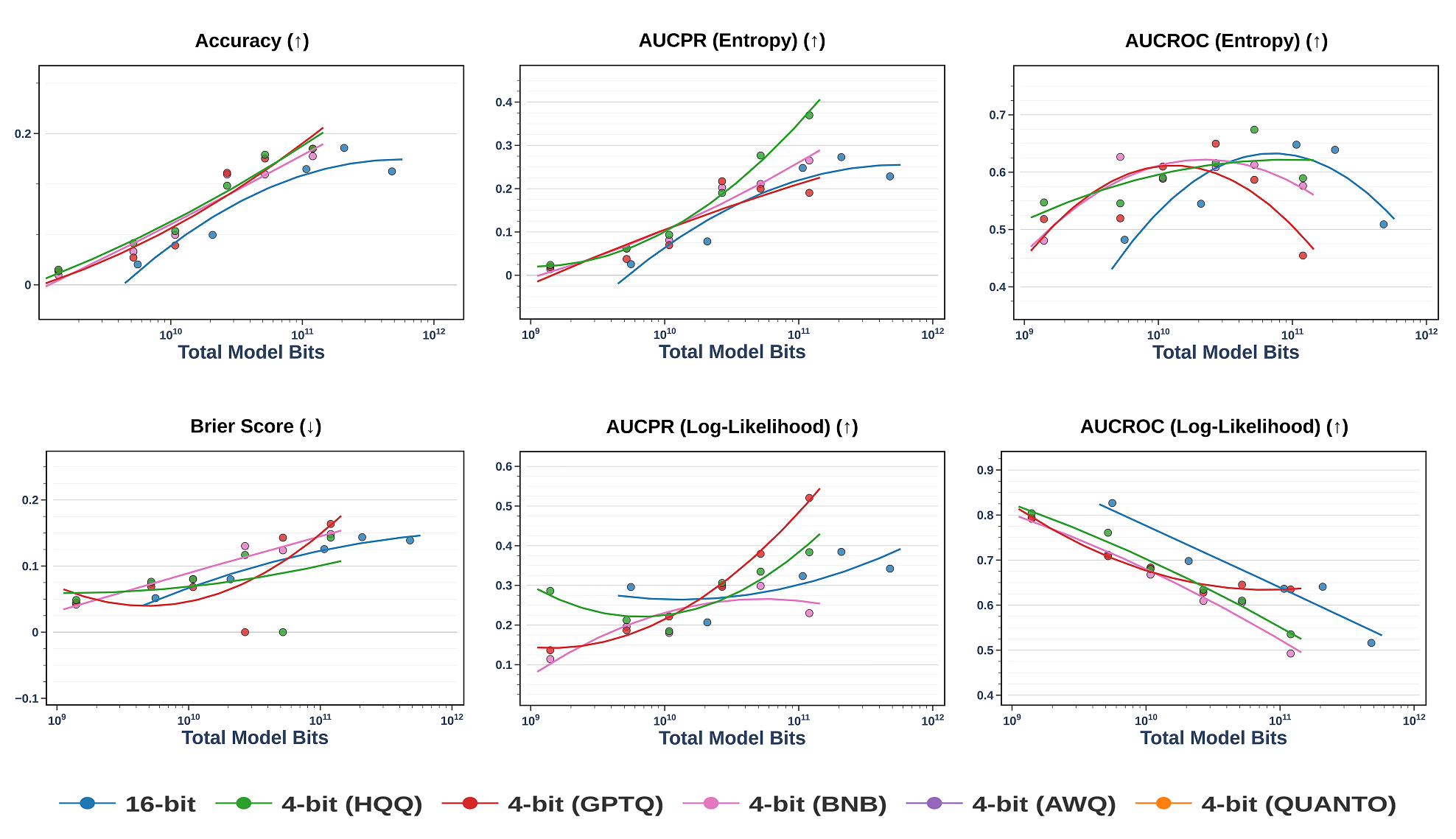}
    \caption{{Bit-level inference scaling trends of OPT models under different 4-bit quantization methods.}} 
    \label{fig:opt_4bit}
    \vspace{-1em}
\end{figure}
\begin{figure}[t]
 \vspace{-0.5em}
    \centering
    \includegraphics[width=0.8\textwidth]{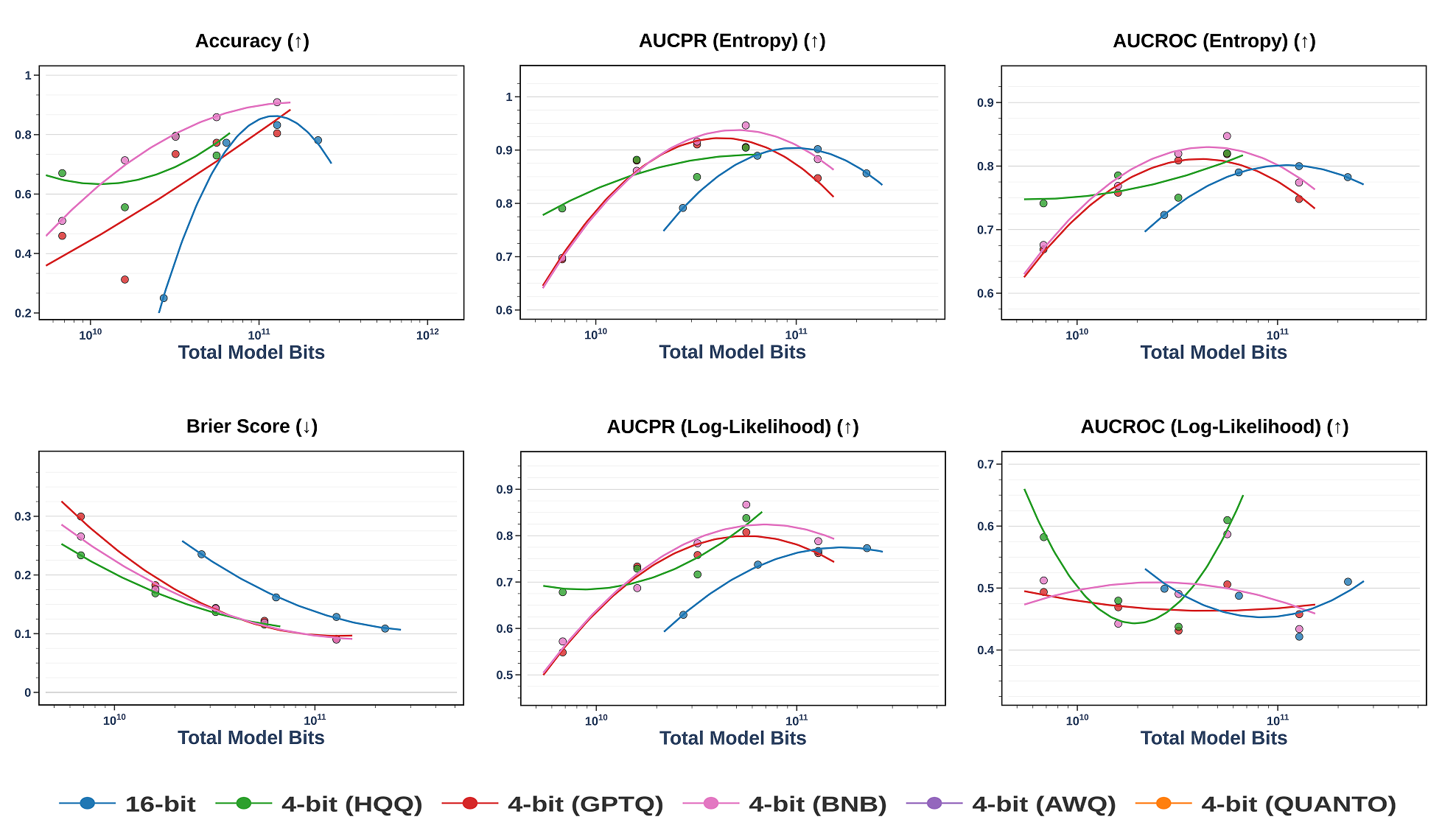}
    \caption{{Bit-level inference scaling trends of Qwen3 models under different 4-bit quantization methods.}} 
    \label{fig:qwen_4bit}
    \vspace{-1em}
\end{figure}

\section{{Additional KLD analysis}}
\begin{figure}[]
 \vspace{-0.5em}
    \centering
    \includegraphics[width=0.8\textwidth]{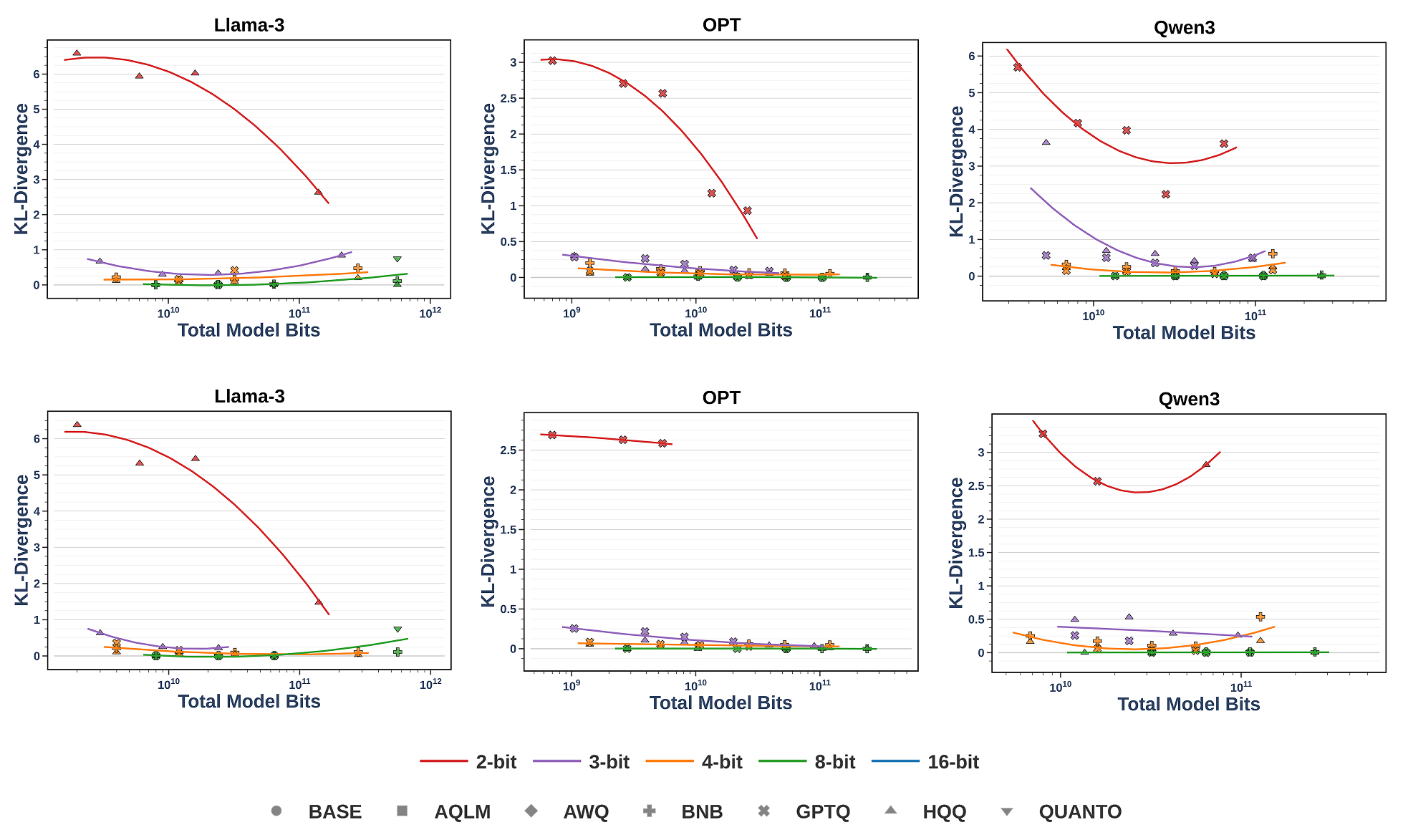}
        \caption{{Additional KL-divergence analysis across LLaMA-3, OPT, and Qwen3 models on Wikitext (top) and C4 (bottom) datasets.}} 
    \label{fig:additional_kld}
    \vspace{-1em}
\end{figure}

{
To further analyze the behavioral shift induced by quantization, we extend the KL-divergence analysis beyond the LLaMA family. In particular, we compare the token-level predictive distributions of quantized models against their corresponding full-precision base models across three model families: LLaMA-3, OPT, and Qwen3. Please refer to \cref{fig:additional_kld}. We evaluate multiple quantization precisions, including 2-, 3-, 4-, and 8-bit models, on two different benchmarks. This allows us to assess whether the observed relationship between quantization precision and behavioral shift is consistent across architectures and tasks.
}

\end{document}